\documentclass[twoside]{article}
\usepackage{preprint2col}
\usepackage{amsmath}   
\usepackage{multirow}
\usepackage{graphicx}
\usepackage{booktabs}
\usepackage{amssymb}
\usepackage{xcolor}

\usepackage[round]{natbib}

\begin{document}

% \twocolumn[

% \aistatstitle{AdaCast: Conditional Parameter Generation for Adaptive Time Series Forecasting}

% \aistatsauthor{ Author 1 \And Author 2 \And  Author 3 }

% \aistatsaddress{ Institution 1 \And  Institution 2 \And Institution 3 } ]

\twocolumn[
\preprinttitle{AdaCast: Conditional Parameter Generation for Adaptive Time Series Forecasting}
\preprintauthor{ Darahaas Nallagatla \And Darryl Cherian Jacob \And Pan He }
\preprintaddress{ Auburn University \And Auburn University \And Auburn University }
]

\begin{abstract}
Time-series foundation models (TSFMs) have achieved strong forecasting performance across domains. However, most adaptation methods remain static. Existing all-in-one methods learn a single set of dataset-level parameter updates and apply the same adapted model to every input. As a result, they cannot adapt the model parameters to the temporal patterns, seasonality and dynamics of each input time series. This limits their ability to produce forecasts that are tailored to heterogeneous inputs. To address this limitation, we propose AdaCast, a conditional parameter generation framework for time-series forecasting.  AdaCast uses a generator to produce input-specific low-rank parameter updates for a frozen pretrained TSFM. These updates adapt the model to each input during both training and inference. Across six public benchmarks, AdaCast consistently outperforms static adaptation baseline in in-domain forecasting and improves  zero-shot generalization to held-out datasets across domains. These results demonstrate that conditional parameter generation provides an effective approach for adaptive forecasting.
\end{abstract}

\begin{figure*}
    % \vspace{.3in}
	\centering
	\includegraphics[width=.98\linewidth]{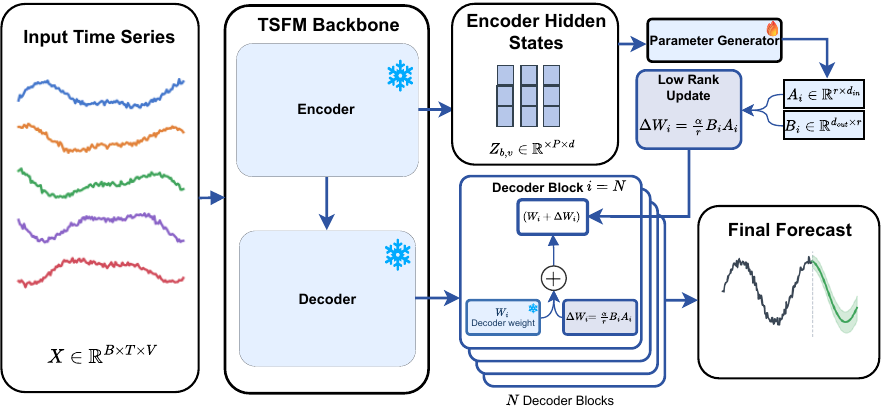}
    % \vspace{.3in}
	\caption{Overview of the \textbf{AdaCast} architecture. Each univariate series $\mathbf{X}_{b,v}$ in the input batch is passed through the frozen backbone encoder, whose hidden states $\mathbf{Z}_{b,v}$ serve as the conditioning representation. The parameter generator takes this representation as its condition and emits a pair of low-rank matrices for every targeted module in every decoder block. Each pair forms the low-rank update for that module, and together they constitute the instance-specific update. These updates are injected additively into the corresponding frozen decoder weights, leaving the pretrained parameters unchanged and yielding the instance-adapted backbone, which produces the final forecast.}
	\label{framework}
    % \vspace{.3in}
\end{figure*}
\section{Introduction}

\label{sec:intro}
Time-series foundation models (TSFMs) have supported broad real-world applications, including
energy demand prediction, traffic management, financial forecasting, and health
monitoring~\citep{survey_of_ts, alonso2026pretrainedtimeseriesfoundationmodels, yanespulido2026timeseriesfoundationmodels}. However, the heterogeneity of such time-series data—spanning diverse temporal dynamics, domain-specific seasonal patterns, and complex inter-variable dependencies—continues to pose a fundamental challenge to developing robust and generalizable forecasting methods~\citep{Wang24deep}.
 
Recent years have seen a major shift from task-specific forecasting models to large foundation models pretrained on massive and heterogeneous time-series datasets. Models such as Chronos~\citep{Ansari24}, TimesFM~\citep{Das24}, and Moirai~\citep{Woo24} show that a single pretrained backbone can support zero-shot forecasting and efficient fine-tuning across diverse domains. This shift reduces the need for domain-specific model design and enables more unified and transferable time-series representations.

In essence, all-in-one TSFM approaches learn a single set of dataset-level parameter updates and apply the same adapted model to every input. However, heterogeneous time series may require different forecasting behaviors, causing optimization to favor shared patterns while underrepresenting domain- and instance-specific features~\citep{Ha17,HyWu26,jacob2026copra}. This issue may be amplified under distribution shift, where test inputs may require different temporal filters, dependency structures, or extrapolation strategies compared to training inputs. The key limitation is therefore not only model capacity, but also the lack of a mechanism that adapts model parameters to each input under distribution shift.
 
To address this limitation, we propose \textit{AdaCast}, a conditional parameter generation framework for time-series forecasting. Unlike existing all-in-one TSFM methods~\citep{Ansari24, Das24,Ekambaram24, Goswami24, Woo24}, AdaCast derives an input-specific feature and uses it to generate lightweight parameter updates for a frozen TSFM backbone. This design allows the model to adjust its forecasting behavior on the fly to the dynamics of each input. Inspired by functional memory~\citep{HyWu26,jacob2026copra}, AdaCast replaces a single compromise update with input-conditioned updates and extends hypernetwork-based adaptation from localized forecasting modules to scalable, cross-domain adaptation of TSFMs.

\subsection{Contributions}
Our contributions are summarized as follows:

\begin{enumerate}

\item \textit{Input-conditioned TSFM adaptation under distribution shift.}
Existing TSFM adaptation methods typically learn a single set of parameter updates and apply the same adapted model to heterogeneous inputs. We instead formulate TSFM adaptation as an input-conditioned problem, where parameter updates are generated for each time series according to its temporal characteristics. This formulation allows the model to better accommodate heterogeneous inputs and distribution shifts at inference time.

\item \textit{Conditional parameter generation for TSFMs.}
We introduce \textbf{AdaCast}, a conditional parameter generation framework that augments a frozen Chronos-Bolt backbone~\citep{Ansari24} with a lightweight parameter generator. For each input time series, AdaCast derives an input-specific  representation and generates low-rank parameter updates that customize the forecasting model to the current input. This enables per-instance adaptation during both training and inference.

\item \textit{Strong in-domain and cross-domain forecasting.}
We evaluate AdaCast on six benchmarks spanning energy, exchange-rate, and weather forecasting. AdaCast consistently outperforms zero-shot Chronos and static adaptation baselines in in-domain settings and achieves strong zero-shot generalization across domains. These comprehensive experiments demonstrate the effectiveness of AdaCast for adaptive forecasting.

\end{enumerate}

\section{Related Work}

\noindent \textbf{Time Series Forecasting.}
Long-term forecasting has progressed from Transformer-based models with decomposition~\citep{Wu21autoformer} or specialized attention~\citep{Zhou21,Liu22,Zhou22}, to patch-based and linear models~\citep{Nie23,Zeng23}, and more recent convolutional and mixing architectures~\citep{Wu23timesnet,Wang24timemixer}. These models perform well in-domain but are typically trained separately for each dataset and have limited zero-shot transfer. Each new domain therefore requires sufficient training data which is impractical when data are scarce or new series appear frequently.

\noindent \textbf{Time Series Foundation Models.}
TSFMs, such as Chronos and Chronos-Bolt~\citep{Ansari24}, TimesFM~\citep{Das24}, Moirai~\citep{Woo24}, MOMENT~\citep{Goswami24}, and TTM~\citep{Ekambaram24}, pretrain a shared backbone on large heterogeneous corpora for zero-shot forecasting. Domain adaptation typically relies on full fine-tuning or dataset-level LoRA~\citep{Hu22}, which produces a fixed parameter set shared across all inputs. Yet even within a single dataset, individual series can differ substantially; for example, the load profiles of residential and industrial electricity customers follow different daily and weekly patterns. A single adapter must average over this heterogeneity, so it may not fully capture the temporal patterns, seasonality, and dynamics of each individual series.

\noindent \textbf{Instance-Specific Conditioning.}
A single set of weights must serve series with very different trends, seasonalities, and dynamics, so many methods inject instance-specific information at inference time. \emph{Retrieval-based} methods fetch similar historical series or segments and fuse them with the input~\citep{han2025retrieval, ning2025tsrag, zhou2026saraf}. \emph{Multimodal} methods add side information such as text or images~\citep{wu2026aurora,wang2025chattime}. \emph{Covariate-aware} methods condition on exogenous or known-future variables~\citep{ansari2025chronos2univariateuniversalforecasting, khwaja2026toto20timeseries}.
Despite their differences, all of these act on the \emph{input or activation} side: they supply extra context while the network weights stay fixed, so the weights are shared across instances and only the inputs differ. 
They also depend on external information that is often unavailable. This leaves open a complementary route: conditioning the \emph{weights} themselves.

\noindent \textbf{Conditional Parameter Generation.}
Conditioning the weights is the premise of hypernetworks~\citep{Ha17}, which generate model parameters from a conditioning signal. In time series forecasting, prior work has used them to generate forecasting parameters~\citep{Lee22,Savchenko26} or to improve robustness under distribution shift~\citep{Duan23}. However, these methods typically condition on channel identities or distribution-level statistics rather than on the individual input series, and target relatively small, task-specific models. Beyond time series, HY-WU~\citep{HyWu26} applies conditional parameter generation to text-guided image editing, and COPRA~\citep{jacob2026copra} to vision-language-model-based video anomaly detection. These works show that generated parameters can adapt a model to each input, but this has not been explored for TSFMs. AdaCast fills this gap by generating input-specific LoRA updates for a frozen TSFM backbone, using only the input series.

\section{Method}
\subsection{Problem Definition}

Let $\mathbf{X} \in \mathbb{R}^{B \times T \times V}$ denote a batch of multivariate time series, where $B$ is the batch size, $T$ is the input sequence length, and $V$ is the number of variables. Given $\mathbf{X}$, the goal is to predict the next $H$ time steps,
$\mathbf{Y} \in \mathbb{R}^{B \times H \times V}$, where $H$ denotes the forecasting horizon. We use Chronos-Bolt~\citep{Ansari24}, a pretrained T5-based~\citep{2020t5} forecasting model, as the backbone $f_{\theta}$, where $\theta$ denotes its model parameters. Following the channel-independent strategy used by Chronos-Bolt, each variable is forecast independently. 

Specifically, for sample $b \in \{1,\ldots,B\}$ and variable $v \in \{1,\ldots,V\}$, we extract the univariate history $\mathbf{x}_{b,v} = \mathbf{X}_{b,:,v} \in \mathbb{R}^{T}$.
The backbone produces forecasts for $Q$ quantile levels over the next $H$ steps:
\begin{equation}
    \hat{\mathbf{q}}_{b,v}
    = f_{\theta}(\mathbf{x}_{b,v})
    \in \mathbb{R}^{Q \times H},
    \label{eq:univariate_forecast}
\end{equation}
where $Q$ is the number of predicted quantiles. Stacking the forecasts over all samples and variables gives
$\hat{\mathbf{Q}} \in \mathbb{R}^{B \times Q \times H \times V}$.
For point forecasting, we use the median ($0.5$ quantile) forecast~\citep{Ansari24, ning2025tsrag, Woo24}, yielding
$\hat{\mathbf{Y}} \in \mathbb{R}^{B \times H \times V}$.

\subsection{Overview}

AdaCast adapts a pretrained TSFM on a per-instance basis, overcoming a key limitation of static adaptation methods that apply the same learned update to all inputs. Given the   univariate input series $\mathbf{x}_{b,v}$  and  a pretrained forecasting backbone $f_\theta$. A conventional static adaptation method learns a single global parameter update $\Delta\theta_{\mathrm{static}}$ and predicts
\begin{equation}
    \hat{\mathbf{q}}_{b,v}
    =
    f_{\theta+\Delta\theta_{\mathrm{static}}}
    \!\left(\mathbf{x}_{b,v}\right).
    \label{eq:static}
\end{equation}
After training, $\Delta\theta_{\mathrm{static}}$ is fixed and shared across all input series. During training, this update is learned from gradients produced by many time series. These series can differ in sampling frequency, seasonality, temporal dynamics, and statistical properties. Therefore, the learned update must capture diverse patterns that are useful across heterogeneous inputs. This can produce a general compromise rather than an update that is specialized for each input series.

Instead, AdaCast replaces the shared global update with an input-conditioned update. For each univariate input series, it computes
\begin{equation}
    \Delta\theta_{b,v}
    =
    \mathcal{G}_{\phi}
    \!\left(
        \operatorname{Enc}_{\theta}
        \!\left(\mathbf{x}_{b,v}\right)
    \right),
    \qquad
    \hat{\mathbf{q}}_{b,v}
    =
    f_{\theta+\Delta\theta_{b,v}}
    \!\left(\mathbf{x}_{b,v}\right),
    \label{eq:adacast}
\end{equation}
where $\operatorname{Enc}_{\theta}$ is the frozen encoder of the pretrained backbone. The function $\mathcal{G}_{\phi}$ is a parameter generator with trainable parameters $\phi$. The encoder maps $\mathbf{x}_{b,v}$ to a latent representation of the input series. The generator then uses this representation to produce the input-specific update $\Delta\theta_{b,v}$.

Unlike the global update in Eq.~\eqref{eq:static}, $\Delta\theta_{b,v}$ is specific to each input series. This allows AdaCast to adapt the backbone to the characteristics of each input. The pretrained backbone parameters $\theta$ remain frozen during training, and only the generator parameters $\phi$ are optimized. This design keeps the adaptation lightweight while allowing the effective model parameters to vary across inputs.

\subsection{Conditional Parameter Generation}

The parameter generator $\mathcal{G}_{\phi}$ is a Transformer that maps the
conditioning sequence
\(
\mathbf{Z}_{b,v}
=
\operatorname{Enc}_{\theta}
\!\left(\mathbf{x}_{b,v}\right)
\)
to input-specific LoRA parameters for the frozen decoder. Following
\citep{HyWu26,jacob2026copra}, the generator represents the LoRA parameters
of each adapted module as a set of latent parameter tokens. The LoRA
matrices are partitioned along the channel dimension into local segments.
Each token has shape $r \times d_{seg}$, where $r$ is the LoRA rank and $d_{seg}$
denotes the size of a local channel segment. Thus, each token represents a
local part of the LoRA parameters while retaining all $r$ rank components.
The tokens are organized by decoder block, adapted module, and channel
segment. This organization preserves the structural location of each token
in the backbone.

The initial parameter tokens are learnable and shared across all input
series. They therefore provide a common structural template for parameter
generation. For each input $\mathbf{x}_{b,v}$, the conditioning sequence
$\mathbf{Z}_{b,v}$ modifies this shared template. In each generator block,
the parameter tokens cross-attend to $\mathbf{Z}_{b,v}$. This operation
incorporates information from the current input into the parameter tokens.
As the tokens pass through the generator blocks, they are progressively
refined into an input-specific parameter representation.

Following~\citep{HyWu26,jacob2026copra}, the generator applies self-attention in two
factorized passes. Intra-layer attention connects parameter tokens from
adapted modules within the same decoder block. Inter-layer attention
connects structurally corresponding tokens across decoder blocks.
Together, these passes model parameter dependencies within and across
decoder layers.
After the final generator block, the refined tokens are separated according to whether they belong to the LoRA $\mathbf{A}$ or $\mathbf{B}$ matrices. Separate projection heads map the corresponding tokens to parameter values. The projected tokens are then detokenized and assembled into the module-specific $\mathbf{A}$ and $\mathbf{B}$ matrices.

% After the final generator block, separate output heads map the refined
% tokens to the two LoRA factors. The resulting values are detokenized and
% assembled into module-specific LoRA matrices. Different decoder modules can
% have different input and output dimensions. The rank-anchored tokenization
% scheme of~\citep{HyWu26} handles these heterogeneous dimensions by
% preserving the full LoRA rank dimension while partitioning the channel
% dimension into local segments. This structured representation allows the
% same generator to produce LoRA parameters for modules with different
% matrix shapes.

\subsection{LoRA Injection and Initialization}

We apply input-specific LoRA updates to the same set of projection modules
$\mathcal{M}$ in all $N$ frozen decoder blocks. For block
$i\in\{1,\ldots,N\}$ and module $m\in\mathcal{M}$, let
$\mathbf{W}^{(i,m)}
\in\mathbb{R}^{d_{\mathrm{out}}^{(i,m)}
\times d_{\mathrm{in}}^{(i,m)}}$
denote the frozen weight matrix. Here, $d_{\mathrm{in}}^{(i,m)}$ and
$d_{\mathrm{out}}^{(i,m)}$ are the module's input and output dimensions. For each input series $\mathbf{x}_{b,v}$, the generator produces
\[
    \mathbf{A}_{b,v}^{(i,m)}
    \in\mathbb{R}^{r\times d_{\mathrm{in}}^{(i,m)}},
    \qquad
    \mathbf{B}_{b,v}^{(i,m)}
    \in\mathbb{R}^{d_{\mathrm{out}}^{(i,m)}\times r},
\]
where $b$ indexes batch samples, $v$ indexes variables, and $r$ is the
LoRA rank. The weight update and effective weight are
\begin{align}
    \Delta\mathbf{W}_{b,v}^{(i,m)}
    &=
    \frac{\alpha}{r}\,
    \mathbf{B}_{b,v}^{(i,m)}
    \mathbf{A}_{b,v}^{(i,m)},
    \label{eq:lora_update}\\
    \widetilde{\mathbf{W}}_{b,v}^{(i,m)}
    &=
    \mathbf{W}^{(i,m)}
    +
    \Delta\mathbf{W}_{b,v}^{(i,m)}.
    \label{eq:lora_effective_weight}
\end{align}

Following LoRA~\citep{Hu22}, we
initialize the weights and any bias of the $\mathbf{B}$ output head to
zero. The initial update is therefore zero, preserving the backbone's
pretrained forecasting behavior. With nonzero $\mathbf{A}$ factors, the
$\mathbf{B}$ head can receive gradients at initialization. As its weights
become nonzero, gradients can also reach earlier generator layers.

The generated updates are used only for the current forward pass and
cleared afterward. We optimize only the generator parameters $\phi$;
all pretrained backbone parameters remain frozen.

\begin{table*}[t]
\centering
\caption{Zero-shot forecasting ($T{=}512$, $H{=}64$).
Best and second-best results are \textbf{bold} and
\underline{underlined}, respectively.
``---'' indicates that zero-shot results are not reported because the
evaluation dataset was included in the backbone's pretraining data
(e.g., Weather). AdaCast achieves the best MSE and MAE on five of six datasets and the
lowest average error, reducing average MSE by 1.0\% over static LoRA
and 2.3\% over the frozen backbone in zero-shot setting.}
\label{tab:main}
\begin{center}
\resizebox{\textwidth}{!}{%
\begin{tabular}{l cc cc cc cc cc cc cc}

\toprule
& \multicolumn{2}{c}{\textbf{AdaCast(Ours)}}
& \multicolumn{2}{c}{Chronos-Bolt\textsubscript{B} LoRA}
& \multicolumn{2}{c}{Chronos-Bolt\textsubscript{B}}
& \multicolumn{2}{c}{MOMENT}
& \multicolumn{2}{c}{TTM\textsubscript{B}}
& \multicolumn{2}{c}{Moirai\textsubscript{B}}
& \multicolumn{2}{c}{TimesFM} \\

\cmidrule(lr){2-3}
\cmidrule(lr){4-5}
\cmidrule(lr){6-7}
\cmidrule(lr){8-9}
\cmidrule(lr){10-11}
\cmidrule(lr){12-13}
\cmidrule(lr){14-15}

Dataset
& MSE & MAE
& MSE & MAE
& MSE & MAE
& MSE & MAE
& MSE & MAE
& MSE & MAE
& MSE & MAE \\
\midrule

ETTh1
& \underline{0.3593} & \underline{0.3635}
& \textbf{0.3569} & \textbf{0.3633}
& 0.3616 & 0.3650
& 0.3920 & 0.4110
& 0.3619 & 0.3710
& 0.3686 & 0.3835
& 0.4254 & 0.3825 \\

ETTh2
& \textbf{0.2450} & \textbf{0.2979}
& \underline{0.2491} & \underline{0.2988}
& 0.2517 & 0.2992
& 0.2742 & 0.3327
& 0.2531 & 0.3032
& 0.2547 & 0.3053
& 0.2894 & 0.3233 \\

ETTm1
& \textbf{0.3017} & \textbf{0.3143}
& \underline{0.3057} & \underline{0.3169}
& 0.3109 & 0.3185
& 0.3506 & 0.3824
& 0.3152 & 0.3248
& 0.5399 & 0.4322
& 0.3321 & 0.3326 \\

ETTm2
& \textbf{0.1485} & \textbf{0.2228}
& 0.1493 & \underline{0.2232}
& \underline{0.1487} & 0.2236
& 0.1703 & 0.2579
& 0.1511 & 0.2405
& 0.1958 & 0.2687
& 0.1703 & 0.2552 \\

Weather
& \textbf{0.1491} & \textbf{0.1789}
& \underline{0.1515} & \underline{0.1816}
& 0.1525 & 0.1825
& 0.1801 & 0.2384
& 0.1543 & 0.1893
& 0.1711 & 0.1912
& --- & --- \\

Exchange
& \textbf{0.0594} & \textbf{0.1683}
& \underline{0.0633} & 0.1732
& 0.0673 & 0.1780
& 0.0979 & 0.2059
& 0.0657 & 0.1725
& 0.0663 & \underline{0.1720}
& 0.0695 & 0.1802 \\

\midrule

Average
& \textbf{0.2105} & \textbf{0.2576}
& \underline{0.2126} & \underline{0.2595}
& 0.2155 & 0.2611
& 0.2442 & 0.3047
& 0.2169 & 0.2669
& 0.2661 & 0.2922
& --- & --- \\

\bottomrule

\end{tabular}%
}
\end{center}
\end{table*}
\begin{figure*}[t]
    \centering
    \includegraphics[width=0.48\textwidth]
    {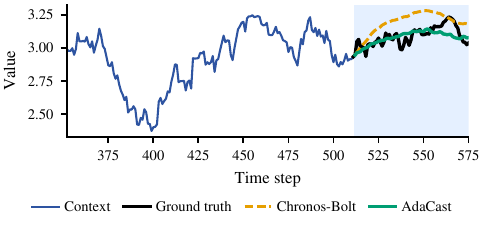}
    \hfill
    \includegraphics[width=0.48\textwidth]
    {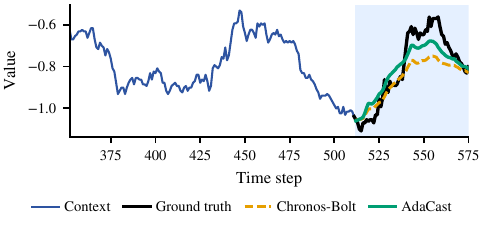}
    \caption{
        Qualitative forecasting examples with context length $T=512$
        and forecast horizon $H=64$.
        Each panel shows the observed context, ground-truth future,
        frozen Chronos-Bolt\textsubscript{B} forecast, and AdaCast
        forecast.
    }
    \label{fig:qualitative_main}
\end{figure*}

\subsection{End-to-End Training}

We train AdaCast end-to-end by optimizing only the generator parameters
$\phi$. All backbone parameters $\theta$, including those of the
conditioning encoder, remain frozen. For each input series $\mathbf{x}_{b,v}$, the adapted backbone predicts
$Q$ quantiles over a forecast horizon of $H$ steps:
\begin{equation}
    \hat{\mathbf{q}}_{b,v}
    =
    f_{\theta+\Delta\theta_{b,v}}
    \!\left(\mathbf{x}_{b,v}\right)
    \in \mathbb{R}^{Q\times H},
    \label{eq:forecast}
\end{equation}
where $\Delta\theta_{b,v}$ is the input-specific update produced by the
generator. The entry $\hat{q}_{b,v}^{(j,h)}$ denotes the predicted value
at quantile level $q_j\in(0,1)$ and forecast step $h$.

For quantile level $q_j$, target value $y$, and predicted value $\hat{q}$,
we use the scaled quantile loss
\begin{equation}
    \ell_{q_j}\!\left(y,\hat{q}\right)
    =
    \begin{cases}
        2q_j\,(y-\hat{q}) & \text{if } y>\hat{q},\\[3pt]
        2(1-q_j)\,(\hat{q}-y) & \text{otherwise.}
    \end{cases}
    \label{eq:pinball_cases}
\end{equation}
The training objective averages over $B$ batch samples, $V$ variables,
and $Q$ quantile levels as:
\begin{equation}
    \mathcal{L}
    =
    \frac{1}{BVQ}
    \sum_{b=1}^{B}
    \sum_{v=1}^{V}
    \sum_{h=1}^{H}
    \sum_{j=1}^{Q}
    \ell_{q_j}\!\left(
        \mathbf{Y}_{b,h,v},
        \hat{q}_{b,v}^{(j,h)}
    \right),
    \label{eq:loss}
\end{equation}
where $\mathbf{Y}_{b,h,v}$ is the target value for sample $b$, variable
$v$, and forecast step $h$. Gradients from $\mathcal{L}$ propagate
through the adapted decoder and generated LoRA factors to update $\phi$.

\section{Experiments}
\label{sec:experiments}

Our experiments address four questions.
\textbf{(1) Zero-shot accuracy:} Does AdaCast improve forecasts on
datasets excluded from generator pretraining?
\textbf{(2) Input-specific conditioning:} Does conditional parameter
generation outperform static LoRA, and do the gains depend on matching
each input with its generated update?
\textbf{(3) Design sensitivity:} Which decoder segments contribute to
adaptation, and how does the LoRA rank affect performance?
\textbf{(4) Backbone transfer:} Do the benefits extend beyond
Chronos-Bolt to other backbones?
We first describe the experimental setup, then examine the questions
through baseline comparisons, controlled interventions, and sensitivity
analyses across multiple datasets.

\subsection{Experimental Setup}
\label{subsec:exp_setup}

\paragraph{Training and evaluation settings.}
For \emph{zero-shot} evaluation, we train the generator on an external
corpus without using target datasets. For \emph{in-domain} evaluation,
we train it on each target dataset's training split. In both settings,
the backbone remains frozen, and the generator produces input-specific
updates from the context at inference.

\paragraph{Datasets.}
We build the generator pretraining corpus from
GiftEvalPretrain~\citep{aksu2024giftevalbenchmarkgeneraltime}, excluding
ETT, Weather, and Exchange. Each variable is treated as an independent
univariate series. We extract approximately five million windows with
$T=512$ context observations and $H=64$ target observations, then
globally shuffle them.

We evaluate on six benchmarks that cover different domains and
sampling intervals.
The four ETT datasets record electricity transformer temperatures
over two years: ETTh1 and ETTh2 use hourly observations, whereas
ETTm1 and ETTm2 use 15-minute observations~\citep{Zhou21}.
Weather contains 21 meteorological indicators recorded at
a German weather station every 10 minutes over one
year~\citep{Wu21autoformer}.
Exchange contains daily exchange rates for eight countries
from 1990 to 2016~\citep{Lai18}.
We use the standard chronological training, validation, and test
splits~\citep{Nie23}.
All experiments use context length $T=512$ and forecast horizon $H=64$.

\paragraph{Evaluation metrics.}
We report mean squared error (MSE) and mean absolute error (MAE),
averaged across variables. For quantile forecasts, we use the predicted
median ($0.5$ quantile). Lower values are better.
% When reporting an average across datasets, we assign equal weight
% to each dataset.

\paragraph{Baselines.}
We compare AdaCast with five pretrained TSFMs:
Chronos-Bolt\textsubscript{B}~\citep{Ansari24},
MOMENT~\citep{Goswami24},
TTM\textsubscript{B}~\citep{Ekambaram24},
Moirai\textsubscript{B}~\citep{Woo24},
and TimesFM~\citep{Das24}.
Their zero-shot results are taken from~\citep{ning2025tsrag}
under the same evaluation protocol.
Frozen Chronos-Bolt\textsubscript{B} is the direct backbone reference:
it forecasts without generated parameter updates.

We also train a static LoRA baseline on the same GiftEvalPretrain
corpus used for AdaCast.
This baseline learns one shared adapter for the frozen
Chronos-Bolt\textsubscript{B} backbone.
The comparison tests whether conditional generation provides gains
beyond adaptation on the same external data.
For the in-domain comparison, we train static LoRA separately on
each target dataset's training split.

\paragraph{Implementation details.}
AdaCast uses the frozen \texttt{amazon/chronos-bolt-base} backbone.
Only the parameter generator $\mathcal{G}_{\phi}$ is optimized.
The generator contains $N_{\mathrm{pg}}$ Transformer blocks with
hidden dimension $d_{\mathrm{pg}}=768$.
We use $N_{\mathrm{pg}}=8$ for zero-shot experiments and
$N_{\mathrm{pg}}=1$ for in-domain experiments.
Unless otherwise stated, the LoRA rank is $r=32$ and the scaling
parameter is $\alpha=128$.
We adapt the same target projection modules in all $N=12$ decoder
blocks. We train with AdamW using learning rate $\eta=10^{-7}$ and batch
size $B=32$.
In-domain hyperparameters are selected on the validation split.
Unless otherwise stated, experiments use a random seed of 1.
The zero-shot input--update pairing experiment reports averages over
seeds $\{1,43,2021\}$.
Experiments run on four NVIDIA RTX PRO 6000 GPUs.
\begin{table}[t]
\caption{
    In-domain conditional generation versus static adaptation.
    All methods use Chronos-Bolt\textsubscript{B}.
    Static LoRA and AdaCast use $r=32$ and $\alpha=128$;
    AdaCast uses $N_{\mathrm{pg}}=1$.
    Each cell reports MSE\,/\,MAE.
    \textbf{Bold} marks the best result. At the same rank and scaling, AdaCast outperforms both the frozen
backbone and static LoRA.
% on every dataset and metric, reduc.ing average
% MSE by 8.8\% and 3.6\%, respectively..
}

\label{tab:abl_conditioning}
\begin{center}
\resizebox{\columnwidth}{!}{%
\begin{tabular}{lccc}
\toprule
\textbf{Dataset}
& \textbf{Frozen Backbone}
& \textbf{Static LoRA}
& \textbf{AdaCast} \\
\midrule
ETTh1
& 0.3616\,/\,0.3650
& 0.3505\,/\,0.3610
& \textbf{0.3431\,/\,0.3594} \\
ETTh2
& 0.2517\,/\,0.2992
& 0.2412\,/\,0.2997
& \textbf{0.2370\,/\,0.2919} \\
ETTm1
& 0.3109\,/\,0.3185
& 0.2840\,/\,0.3155
& \textbf{0.2603\,/\,0.3040} \\
ETTm2
& 0.1487\,/\,0.2236
& 0.1401\,/\,0.2171
& \textbf{0.1385\,/\,0.2160} \\
\bottomrule
\end{tabular}%
}

\end{center}
\end{table}

\subsection{Zero-Shot Forecasting Results}
\label{subsec:main_results}

We first test whether generator pretraining improves forecasting
on datasets excluded from its training corpus.
Table~\ref{tab:main} compares AdaCast with other  TSFMs
and the corpus-trained static LoRA baseline.

AdaCast achieves the lowest MSE and MAE on five of the six datasets:
ETTh2, ETTm1, ETTm2, Weather, and Exchange.
It also achieves the lowest average errors, with MSE of 0.2105
and MAE of 0.2576.
Compared with the frozen pretrained baselines, AdaCast improves
both metrics on all six datasets.
Relative to its own frozen backbone,
Chronos-Bolt\textsubscript{B}, it reduces average MSE by 2.3\%.
Figure~\ref{fig:qualitative_main} provides example forecasts for
comparison with the frozen backbone.

\subsection{Input-specific Conditioning}
\label{subsec:conditioning}

\noindent \textbf{Instance-conditioned versus static adaptation}. We next test whether input-conditioned adaptation improves in-domain
forecasting over a single shared LoRA adapter. We compare three models
on the four ETT datasets: frozen Chronos-Bolt\textsubscript{B},
in-domain static LoRA, and in-domain AdaCast. Static LoRA learns one fixed pair of factors
$(\mathbf{A},\mathbf{B})$ per target module and applies it to all inputs.
AdaCast instead generates these factors from each input series'
representation during both training and inference. This allows the
updates to reflect the temporal characteristics of individual series
rather than remain fixed.

Both methods are trained on each target dataset's training split using
the same frozen backbone. They adapt the same projection modules in
all 12 decoder blocks, with LoRA rank $r=32$ and scaling parameter
$\alpha=128$. Static LoRA learns shared updates, whereas AdaCast
generates input-specific updates. AdaCast uses a one-block generator
($N_{\mathrm{pg}}=1$) with learning rate $\eta=10^{-7}$.

% Both adapted models are trained on each target dataset's training split.
% They use the same frozen backbone and target the same projection modules
% in all 12 decoder blocks. Both use LoRA rank $r=32$ and scaling parameter
% $\alpha=128$. These settings match the structure of the applied updates,
% while the methods differ in how those updates are learned and generated.
% AdaCast uses a one-block parameter generator ($N_{\mathrm{pg}}=1$)
% with learning rate $\eta=10^{-7}$.

\begin{figure}[t]
\centering
\includegraphics[width=\columnwidth]
{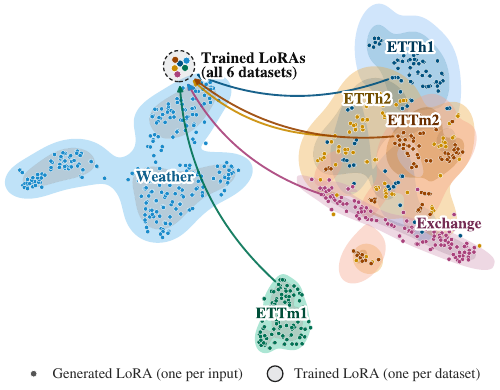}
\caption{
    Joint t-SNE visualization of  LoRA parameters
    across six datasets.
    Static adaptation contributes one adapter per dataset;
    AdaCast contributes one adapter per sampled input.   \textbf{Generated LoRAs} cover a wide region of
    parameter space and are organized by dataset: each input receives its own
    update. \textbf{Trained LoRAs} are fixed points: every
    input from a dataset receives the same update.
    Each arrow marks the \emph{LoRA gap} between a dataset's trained LoRA and
    the generated LoRAs for the same dataset.
}
\label{fig:lora_tsne}
\end{figure}

Table~\ref{tab:abl_conditioning} shows that AdaCast achieves lower
MSE and MAE than both static LoRA and the frozen backbone on all
four datasets.
The largest MSE reduction over static LoRA occurs on ETTm1. The same trend holds on Weather and Exchange, where AdaCast
again outperforms static LoRA on both metrics
(Table~6 in Appendix).
These results support conditional generation under matched LoRA
configurations.

However, AdaCast introduces a trainable generator.
The next experiment therefore tests the value of input-specific
pairing without changing the trained model.

Figure~\ref{fig:lora_tsne} provides a qualitative view of the
adapter parameters.
We flatten six static adapters, one trained per dataset, and
AdaCast adapters generated for inputs from the same datasets.
We then embed the vectors jointly with t-SNE.
The generated adapters form dataset-specific groups in this
projection, whereas the six static adapters appear close together.
% Because the two methods contribute different numbers of points,
% we treat this visualization as illustrative rather than as evidence
% of better adapter quality.

\begin{table}[t]

\centering
\caption{
    Effect of shuffling generated LoRA updates across test inputs.
    Entries report MSE.
    $\Delta$ is shuffled MSE minus matched MSE;
    positive values indicate higher error after shuffling.
    % Zero-shot results are averaged over three seeds. 
    Shuffling increases average error in both settings, suggesting that
AdaCast benefits from updates matched to each input.
% The effect is strongest in-domain by an average MSE $+2.1\%$ vs.\ $+0.6\%$
% zero-shot.
}
\label{tab:shuffle}
\begin{center}
\resizebox{\columnwidth}{!}{%
\begin{tabular}{lccc ccc}
\toprule
& \multicolumn{3}{c}{\textbf{Zero-shot}}
& \multicolumn{3}{c}{\textbf{In-domain}} \\
\cmidrule(lr){2-4}
\cmidrule(lr){5-7}
\textbf{Dataset}
& Matched & Shuffled & $\Delta$
& Matched & Shuffled & $\Delta$ \\
\midrule
ETTh1
& 0.3582 & 0.3594 & \textcolor{red!70!black}{$+0.0012$}
& 0.3431 & 0.3450 & \textcolor{red!70!black}{$+0.0019$} \\
ETTh2
& 0.2442 & 0.2478 & \textcolor{red!70!black}{$+0.0036$}
& 0.2370 & 0.2449 & \textcolor{red!70!black}{$+0.0079$} \\
ETTm1
& 0.3008 & 0.3005 & \textcolor{green!50!black}{$-0.0003$}
& 0.2603 & 0.2666 & \textcolor{red!70!black}{$+0.0063$} \\
ETTm2
& 0.1481 & 0.1494 & \textcolor{red!70!black}{$+0.0013$}
& 0.1385 & 0.1401 & \textcolor{red!70!black}{$+0.0016$} \\
Weather
& 0.1490 & 0.1491 & \textcolor{red!70!black}{$+0.0001$}
& 0.1255 & 0.1295 & \textcolor{red!70!black}{$+0.0040$} \\
Exchange
& 0.0591 & 0.0601 & \textcolor{red!70!black}{$+0.0010$}
& 0.0571 & 0.0602 & \textcolor{red!70!black}{$+0.0031$} \\
\midrule
Average
& 0.2099 & 0.2111 & \textcolor{red!70!black}{$+0.0012$}
& 0.1936 & 0.1977 & \textcolor{red!70!black}{$+0.0041$} \\
\bottomrule
\end{tabular}%
}
\end{center}
\end{table}

\noindent \textbf{Does the input--update pairing matter?}
 We test whether AdaCast benefits from matching each input with its own
generated LoRA update. At inference, we replace each input's update
with one generated for a different, randomly selected input from the
same test set. The trained generator, backbone, and test inputs remain
unchanged; only the input--update pairing changes. If updates are
largely interchangeable within a dataset, shuffling should have little
effect on accuracy. Table~\ref{tab:shuffle} compares matched and
shuffled updates.

In-domain, shuffling increases MSE on all six datasets. Average MSE
rises from 0.1936 to 0.1977, with the largest increases on ETTh2
($+0.0079$) and ETTm1 ($+0.0063$). These results support a benefit
from matching inputs with their generated updates. The zero-shot effect is smaller. Shuffling increases MSE on five
datasets and slightly decreases it on ETTm1. Average MSE rises from
0.2099 to 0.2111, indicating a smaller benefit from correct pairing. 

% Generator depth is fixed within each matched--shuffled comparison
% but differs across settings: $N_{\mathrm{pg}}=8$ for zero-shot and
% $N_{\mathrm{pg}}=1$ for in-domain evaluation. Thus, each comparison
% tests pairing within a fixed model, but the difference in sensitivity
% across settings may reflect both the training regime and generator depth.
% However, shuffling still uses multiple generated
% updates. It does not test whether one shared update would achieve
% the same accuracy.

\subsection{Contributions of Decoder Segments}
\label{sec:abl_depth}

We divide the 12 decoder blocks into three equal segments and disable
LoRA updates in one segment at a time. The other segments remain
adapted, and all backbone weights stay frozen.

% \begin{table}[t]
% \centering
% \caption{
%     In-domain decoder-segment ablation. Segments 1--3 contain blocks
%     1--4, 5--8, and 9--12, respectively.
%     Cells report MSE\,/\,MAE; \textbf{bold} marks the largest error
%     per metric on each dataset.
% }
% \label{tab:abl_segments}
% \resizebox{\columnwidth}{!}{%
% \begin{tabular}{lcccc}
% \toprule
% & \multicolumn{3}{c}{\textbf{Segment Disabled}}
% & \textbf{Full} \\
% \cmidrule(lr){2-4}
% \textbf{Dataset}
% & \textbf{Seg 1}
% & \textbf{Seg 2}
% & \textbf{Seg 3}
% & \textbf{AdaCast} \\
% \midrule
% ETTh1
% & \textbf{0.3510\,/\,0.3617}
% & 0.3460\,/\,0.3596
% & 0.3491\,/\,0.3603
% & 0.3431\,/\,0.3594 \\
% ETTh2
% & \textbf{0.2432\,/\,0.2946}
% & 0.2416\,/\,0.2940
% & 0.2377\,/\,0.2923
% & 0.2370\,/\,0.2919 \\
% ETTm1
% & \textbf{0.2957\,/\,0.3132}
% & 0.2751\,/\,0.3056
% & 0.2747\,/\,0.3064
% & 0.2603\,/\,0.3040 \\
% ETTm2
% & \textbf{0.1447\,/\,0.2202}
% & 0.1400\,/\,0.2168
% & 0.1403\,/\,0.2168
% & 0.1385\,/\,0.2160 \\
% \bottomrule
% \end{tabular}%
% }
% \end{table}

Disabling any segment increases both metrics on all four datasets
(Table~\ref{tab:abl_segments}). Removing early-segment updates causes
the largest degradation. All tested configurations still outperform the frozen
backbone (Table~\ref{tab:abl_conditioning}), indicating that adaptation
remains beneficial with one segment disabled.

% We examine where generated updates contribute most within the decoder.
% We divide the 12 decoder blocks into three contiguous segments:
% early blocks 1--4, middle blocks 5--8, and late blocks 9--12.
% We disable LoRA injection in one segment at a time while retaining
% the updates in the other two segments.
% All pretrained backbone weights remain frozen.

\begin{table}[t]
\centering
\caption{
    Effect of disabling LoRA updates in one decoder segment.
    Segments 1, 2, and 3 contain blocks 1--4, 5--8, and 9--12,
    respectively.
    Results use the in-domain setting.
    Each cell reports MSE\,/\,MAE.
    \textbf{Bold} marks the largest error among the segment-disabled
    configurations. Disabling updates in any segment increases error, suggesting that all
decoder segments contribute to AdaCast's gains.
}
% The early blocks matter most as disabling Segment~1 raises average MSE by
% 5.7\%, more than twice the increase from Segment~2 or Segment~3.

\label{tab:abl_segments}
\begin{center}
\resizebox{\columnwidth}{!}{%
\begin{tabular}{lcccc}
\toprule
& \multicolumn{3}{c}{\textbf{Segment Disabled}}
& \textbf{Full} \\
\cmidrule(lr){2-4}
\textbf{Dataset}
& \textbf{Seg 1}
& \textbf{Seg 2}
& \textbf{Seg 3}
& \textbf{AdaCast} \\
\midrule
ETTh1
& \textbf{0.3510\,/\,0.3617}
& 0.3460\,/\,0.3596
& 0.3491\,/\,0.3603
& 0.3431\,/\,0.3594 \\
ETTh2
& \textbf{0.2432\,/\,0.2946}
& 0.2416\,/\,0.2940
& 0.2377\,/\,0.2923
& 0.2370\,/\,0.2919 \\
ETTm1
& \textbf{0.2957\,/\,0.3132}
& 0.2751\,/\,0.3056
& 0.2747\,/\,0.3064
& 0.2603\,/\,0.3040 \\
ETTm2
& \textbf{0.1447\,/\,0.2202}
& 0.1400\,/\,0.2168
& 0.1403\,/\,0.2168
& 0.1385\,/\,0.2160 \\
\bottomrule
\end{tabular}%
}
\end{center}
\end{table}

% Table~\ref{tab:abl_segments} shows that disabling the early segment
% causes the largest increase in both metrics on all four datasets.
% The largest MSE change occurs on ETTm1, from 0.2603 to 0.2957.
% Disabling either the middle or late segment also increases both
% metrics on every dataset.
% These results indicate that all three segments contribute to the
% full model, with the early segment having the largest effect
% under this intervention. All tested segment-disabled configurations still outperform the
% frozen backbone reported in Table~\ref{tab:abl_conditioning}.

% This experiment does not test whether a single adapted segment
% would be sufficient.

\begin{figure}[t]
\centering
\includegraphics[width=\columnwidth]{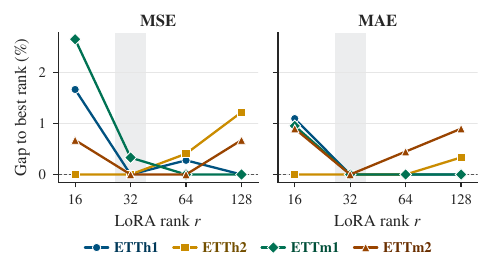}
\caption{
    Sensitivity of zero-shot forecasting to LoRA rank.
Curves show each dataset's relative error gap (\%) to its best tested
rank: MSE (left) and MAE (right). Zero indicates the best result,
including ties. The shaded band marks the default rank $r=32$.
}
\label{fig:abl_rank}
\end{figure}

\subsection{Sensitivity to LoRA Rank}
\label{subsec:rank}

We evaluate LoRA ranks $r\in\{16,32,64,128\}$ in the zero-shot setting.
All configurations use $N_{\mathrm{pg}}=8$, learning rate
$\eta=10^{-7}$, and the same five-million-window pretraining corpus.

Figure~\ref{fig:abl_rank} shows that $r=16$ underperforms compared to the higher
ranks on ETTh1 and ETTm1. Among $r\in\{32,64,128\}$, reported MSE
differences are at most 0.003, with no consistently best rank.
The default $r=32$ matches or nearly matches the lowest MSE on each
dataset while requiring fewer generated parameters than $r=64$
or $r=128$.

\subsection{Extension to a Decoder-Only Backbone}
\label{subsec:backbone}

We apply AdaCast to TimesFM~2.5~\citep{Das24} to test its effectiveness
beyond Chronos-Bolt. Because TimesFM has no separate encoder, we
extract the conditioning sequence from its final decoder hidden states:
$\mathbf{Z}_{b,v}
    =
    f_{\theta}^{(L)}
    \!\left(\mathbf{x}_{b,v}\right)
    \in \mathbb{R}^{P\times d_{\mathrm{tfm}}}$,
where $f_{\theta}^{(L)}$ returns the hidden states of the final decoder
layer, $P$ is the number of context representation tokens, and
$d_{\mathrm{tfm}}$ is the hidden dimension. TimesFM uses causal attention, so each token attends only to the tokens
before it, only after the entire context has been processed does the
final layer hold representations that reflect the full input.
The generator produces
$\Delta\theta_{b,v}=\mathcal{G}_{\phi}(\mathbf{Z}_{b,v})$,
and a second backbone pass applies these updates to produce the forecast.
Inference therefore requires two backbone passes, in addition to  the generator
computation. The second pass is necessary because the generated updates modify
layers that precede the final layer, while $\mathbf{Z}_{b,v}$ is
available only after the full pass completes; the updates therefore
cannot be applied within the pass that produces them.

We evaluate both zero-shot and in-domain generator training using the
settings in Section~\ref{subsec:exp_setup}.
Table~\ref{tab:timesfm_backbone} compares the adapted models with
frozen TimesFM~2.5.

\begin{table}[t]
\centering
\caption{
    AdaCast on TimesFM~2.5 ($T=512$, $H=64$).
    Cells report MSE\,/\,MAE.
    \textbf{Bold} marks the best result for each metric. AdaCast transfers to a second backbone, lowering
MSE over frozen TimesFM~2.5 on every dataset in both settings.
Average MSE drops by 3.8\% in zero-shot and 8.5\% in-domain.
}
\label{tab:timesfm_backbone}
\begin{center}
\resizebox{\columnwidth}{!}{%
\begin{tabular}{lccc}
\toprule
&
& \multicolumn{2}{c}{\textbf{AdaCast--TimesFM 2.5}} \\
\cmidrule(lr){3-4}
\textbf{Dataset}
& \textbf{Frozen TimesFM 2.5}
& \textbf{Zero-shot}
& \textbf{In-domain} \\
\midrule
ETTh1
& 0.3734\,/\,0.3729
& 0.3628\,/\,0.3766
& \textbf{0.3475}\,/\,\textbf{0.3656} \\
ETTh2
& 0.2792\,/\,0.3151
& 0.2610\,/\,0.3132
& \textbf{0.2548}\,/\,\textbf{0.3069} \\
ETTm1
& 0.2988\,/\,0.3186
& 0.2874\,/\,0.3175
& \textbf{0.2764}\,/\,\textbf{0.3150} \\
ETTm2
& 0.1574\,/\,0.2295
& 0.1534\,/\,0.2301
& \textbf{0.1410}\,/\,\textbf{0.2228} \\
Weather
& 0.1430\,/\,\textbf{0.1686}
& 0.1403\,/\,0.1689
& \textbf{0.1289}\,/\,0.1694 \\
Exchange
& 0.0651\,/\,0.1710
& 0.0621\,/\,0.1706
& \textbf{0.0559}\,/\,\textbf{0.1644} \\
\midrule
Average
& 0.2195\,/\,0.2626
& 0.2112\,/\,0.2628
& \textbf{0.2008}\,/\,\textbf{0.2574} \\
\bottomrule
\end{tabular}%
}
\end{center}
\end{table}

Both AdaCast variants reduce MSE on all six datasets.
Average MSE decreases from 0.2195 to 0.2112 with zero-shot adaptation
and 0.2008 with in-domain adaptation.
In-domain adaptation also lowers average MAE from 0.2626 to 0.2574,
whereas zero-shot adaptation slightly increases it to 0.2628.
Thus, AdaCast's MSE gains extend to a second backbone, although
zero-shot MAE gains remain mixed.

\section{Conclusion and Discussion}

We introduced AdaCast, which generates input-conditioned low-rank
updates for a frozen forecasting backbone without test-time optimization.
Experiments on six benchmarks show improved zero-shot forecasting over
frozen Chronos-Bolt, while in-domain comparisons show gains over static
LoRA. The MSE improvements extend to the decoder-only TimesFM~2.5
backbone as well.
Shuffling generated updates increases average error in both settings,
supporting a benefit from matching each input with its own update.
The larger effect in the tested in-domain configuration suggests
stronger reliance on correct pairing. However, differences in generator
depth prevent attributing this contrast solely to the training regime.
Decoder ablations show the greatest sensitivity to removing early-block
updates, while the rank study supports $r=32$ as a practical choice.

Several questions remain. AdaCast is an univariate model,
leaving cross-variate conditioning unexplored. The greater sensitivity
to early-block updates also requires further study. In addition, the
TimesFM implementation incurs an extra backbone pass. Future work will
explore cross-variate conditioning, clarify the role of decoder depth,
strengthen input-specific adaptation on unseen domains, and reduce
inference overhead. Overall, these results support conditional parameter
generation as a promising to shared static adaptation for time-series forecasting.

\bibliography{references}

\clearpage
% \onecolumn
% \thispagestyle{empty}
\appendix
% \preprinttitle{Supplementary Material}

\section{Dataset Statistics}
\label{app:dataset_stats}
We evaluate AdaCast on six publicly available time series forecasting
benchmarks spanning energy, meteorological, and financial domains. Table
\ref{tab:dataset_stats} reports the exact statistics for each dataset,
where $C$ and $T$ denote the number of channels (variates) and total
timesteps, respectively, and $(N_{\mathrm{train}}, N_{\mathrm{val}},
N_{\mathrm{test}})$ denotes the corresponding split sizes.

\begin{table}[h]
\centering
\caption{Statistics of evaluation datasets.}
\label{tab:dataset_stats}
\begin{center}
\resizebox{\columnwidth}{!}{%
\begin{tabular}{l c c c}
\toprule
\textbf{Dataset} & $C$ & $T$ & $(N_{\mathrm{train}}, N_{\mathrm{val}}, N_{\mathrm{test}})$ \\
\midrule
ETTh1    & 7  & 17{,}420 & (8065, 2817, 2817) \\
ETTh2    & 7  & 17{,}420 & (8065, 2817, 2817) \\
ETTm1    & 7  & 69{,}680 & (33985, 11457, 11457) \\
ETTm2    & 7  & 69{,}680 & (33985, 11457, 11457) \\
Exchange Rate & 8  & 7{,}588  & (4736, 697, 1454) \\
Weather  & 21 & 52{,}696 & (36312, 5207, 10476) \\
\bottomrule
\end{tabular}%
}
\end{center}
\end{table}

\begin{table}[t]
\centering
\caption{Training cost of static LoRA and AdaCast on the frozen
Chronos-Bolt backbone, single NVIDIA RTX PRO 6000 ($T{=}512$, $H{=}64$,
$r{=}32$, $\alpha{=}128$). Per-series time is epoch time divided by the
number of univariate series processed per epoch.}
\label{tab:traincost}
\small
\setlength{\tabcolsep}{4pt}
\begin{center}
\begin{tabular}{lrr}
\toprule
& Static LoRA & AdaCast \\
\midrule
Frozen backbone params & 205{,}292{,}928 & 205{,}292{,}928 \\
Trainable params       &   7{,}667{,}712 &  12{,}864{,}000 \\
Trainable fraction     & 3.60\%          & 5.90\% \\
Batch size             & 32              & 32 \\
\midrule
\multicolumn{3}{l}{\textit{Time per univariate series (ms)}} \\
\quad ETTh1   & 0.39 & 12.22 \\
\quad ETTh2   & 0.40 & 12.24 \\
\quad ETTm1   & 0.39 & 12.20 \\
\quad ETTm2   & 0.39 & 12.19 \\
\quad Weather & 0.30 & 12.27 \\
\quad Exchange& 0.39 & 12.07 \\
\midrule
\multicolumn{3}{l}{\textit{Wall-clock per epoch}} \\
\quad ETTh1   & 22\,s   & 690\,s \\
\quad ETTm1   & 93\,s   & 2{,}903\,s \\
\quad Weather & 229\,s  & 8{,}958\,s \\
\bottomrule
\end{tabular}
\end{center}

\end{table}

\begin{table}[t]
\centering
\caption{Inference cost on ETTh1 ($T{=}512$, $H{=}64$, batch size 32,
224 univariate series per batch, single GPU).}
\label{tab:inferencecost}
\small
\begin{center}
\resizebox{\columnwidth}{!}{%
\begin{tabular}{lrrr}
\toprule
& Chronos-Bolt & Static LoRA & AdaCast \\
\midrule
Trainable params      & 0 & 7{,}667{,}712 & 12{,}864{,}000 \\
Time per series (ms)  & 0.21 & 0.23 & 2.60 \\
Throughput (series/s) & 4{,}680 & 4{,}396 & 384 \\
Peak memory (GB)      & 1.49 & 1.52 & 21.6 \\
\bottomrule
\end{tabular}%
}
\end{center}
\end{table}

 \section{Baseline Methods}
\label{app:baselines}
 
We compare AdaCast against six publicly available time series
foundation models, evaluated Zeroshot without any dataset-specific
fine-tuning. These baselines span five distinct architectural families:
tokenized autoregressive transformers, patch-based direct-forecasting
transformers, masked-reconstruction encoder transformers, an MLP-Mixer
model with no self-attention, and a decoder-only causal transformer.
 
\paragraph{Chronos (Base)~\cite{Ansari24}.}
The original Chronos formulation casts forecasting as a language
modeling problem: the input series is scaled and quantized into a fixed
discrete vocabulary, and a T5-based encoder-decoder is trained with a
standard cross-entropy objective over this vocabulary, identically to a
text-based LLM. At inference time, future trajectories are sampled
autoregressively, token by token, from the learned categorical
distribution, and multiple sampled trajectories are aggregated to form
a probabilistic forecast.
 
\paragraph{Chronos-Bolt (Base)~\cite{Ansari24}.}
Chronos-Bolt retains the same T5 encoder-decoder backbone as Chronos
but replaces token-level quantization with patch-based tokenization:
the historical context is chunked into patches, embedded, and passed
through the encoder, after which the decoder directly regresses
multiple quantile levels for the entire forecast horizon in a single
forward pass rather than decoding autoregressively. This direct
multi-step formulation is also the backbone adapted by AdaCast.
 
\paragraph{MOMENT~\cite{Goswami24}.}
MOMENT uses only the encoder stack of a T5-style architecture, pretrained
via a masked-reconstruction objective: the input series is divided into
patches, a random subset of patch embeddings is replaced with a learnable
mask token, and the model is trained to reconstruct the original values
through a lightweight reconstruction head. Since this pretraining
objective is reconstructive rather than forecasting-native, MOMENT is
evaluated through its publicly released forecasting head built on top of
the pretrained encoder, rather than through autoregressive or direct
quantile decoding as in the other baselines.
 
\paragraph{TTM-B~\cite{Ekambaram24}.}
Tiny Time Mixers (TTM) departs from the transformer family entirely,
built instead from lightweight TSMixer blocks that mix information
across the patch, time, and channel dimensions using only MLP layers and
gated attention, with no self-attention. TTM is pretrained on public multi-resolution time series
corpora using adaptive patching and resolution-prefix tuning to remain
extremely lightweight while transferring across sampling frequencies.
 
\paragraph{Moirai (Base)~\cite{Woo24}.}
Moirai is a masked-encoder-only transformer designed for universal
forecasting across arbitrary numbers of variates and sampling
frequencies. It introduces any-variate attention, which flattens a
multivariate series into a single sequence and uses a binary attention
bias together with rotary position embeddings to let the model attend
flexibly across variates without a fixed channel ordering, along with
multi-patch-size input and output projections to support cross-frequency
inputs. The output head parameterizes a mixture distribution over future
values rather than a fixed quantile set.
 
\paragraph{TimesFM~\cite{Das24}.}
TimesFM is a decoder-only transformer using causal self-attention,
where the input context is broken into non-overlapping patches and
processed through a residual MLP block before entering the transformer
stack. Unlike standard autoregressive language models, TimesFM is
trained with an output patch length longer than its input patch length,
allowing it to predict a block of future steps per decoding pass and
substantially reducing the number of autoregressive iterations needed
for long-horizon forecasts.

\section{Evaluation}
\subsection{Multi-Seed Evaluation}
\label{app:seeds}
To assess run-to-run
variability, we repeat pretraining and evaluation with three
seeds, $\{1, 43, 2021\}$, holding all other settings fixed
($N_{\mathrm{pg}} = 8$, $r = 32$, $\alpha = 128$, $\eta = 1e-7$,
and the same 5M-window GiftEvalPretrain corpus). Table~\ref{tab:seeds}
reports the mean and standard deviation over the three runs, alongside the
frozen Chronos-Bolt$_{B}$ backbone, which is deterministic at inference and
therefore has no seed variation.

\begin{table}[t]
\centering
\caption{Multi-seed Zeroshot forecasting results ($T{=}512$, $H{=}64$).
AdaCast entries are mean $\pm$ standard deviation over seeds
$\{1, 43, 2021\}$. Each cell
reports MSE and MAE.}
\small
\setlength{\tabcolsep}{4pt}
\begin{center}
\resizebox{\columnwidth}{!}{%
\begin{tabular}{lcccc}
\toprule
& \multicolumn{2}{c}{AdaCast (Zero-shot)} & \multicolumn{2}{c}{Chronos-Bolt$_{B}$} \\
\cmidrule(lr){2-3}\cmidrule(lr){4-5}
Dataset & MSE & MAE & MSE & MAE \\
\midrule
ETTh1    & $0.3582 \pm 0.0010$ & $0.3635 \pm 0.0002$ & 0.3616 & 0.3650 \\
ETTh2    & $0.2442 \pm 0.0009$ & $0.2968 \pm 0.0010$ & 0.2517 & 0.2992 \\
ETTm1    & $0.3008 \pm 0.0011$ & $0.3134 \pm 0.0009$ & 0.3109 & 0.3185 \\
ETTm2    & $0.1481 \pm 0.0005$ & $0.2226 \pm 0.0002$ & 0.1487 & 0.2236 \\
Weather  & $0.1490 \pm 0.0002$ & $0.1773 \pm 0.0019$ & 0.1525 & 0.1825 \\
Exchange & $0.0591 \pm 0.0003$ & $0.1675 \pm 0.0008$ & 0.0673 & 0.1780 \\
\midrule
Average  & $0.2099$ & $0.2569$ & 0.2155 & 0.2611 \\
\bottomrule
\end{tabular}%
}
\end{center}

\label{tab:seeds}
\end{table}

\begin{table*}[]
\centering
\caption{Effect of generator depth $N_{\mathrm{pg}}$ on Zeroshot performance
($T{=}512$, $H{=}64$, $r{=}32$). Each cell reports MSE\,/\,MAE.
$N_{\mathrm{pg}} = 8$ is the configuration used for all Zeroshot results in
the main paper.}
\label{tab:pgdepth}
\small
\setlength{\tabcolsep}{4pt}
\begin{center}
\begin{tabular}{lcccc}
\toprule
Dataset & $N_{\mathrm{pg}}{=}1$ & $N_{\mathrm{pg}}{=}4$ & $N_{\mathrm{pg}}{=}8$ & $N_{\mathrm{pg}}{=}16$ \\
\midrule
ETTh1    & 0.3602 / 0.3634 & 0.3601 / 0.3638 & 0.3593 / 0.3635 & 0.3598 / 0.3638 \\
ETTh2    & 0.2444 / 0.2972 & 0.2452 / 0.2981 & 0.2450 / 0.2979 & 0.2450 / 0.2977 \\
ETTm1    & 0.3021 / 0.3145 & 0.3025 / 0.3149 & 0.3017 / 0.3143 & 0.3024 / 0.3144 \\
ETTm2    & 0.1484 / 0.2239 & 0.1483 / 0.2244 & 0.1485 / 0.2228 & 0.1490 / 0.2247 \\
Weather  & 0.1501 / 0.1891 & 0.1497 / 0.1824 & 0.1491 / 0.1789 & 0.1492 / 0.1785 \\
Exchange & 0.0654 / 0.1701 & 0.0644 / 0.1697 & 0.0594 / 0.1683 & 0.0591 / 0.1683 \\
\midrule
Average  & 0.2118 / 0.2597 & 0.2117 / 0.2589 & 0.2105 / 0.2576 & 0.2108 / 0.2579 \\
\bottomrule
\end{tabular}
\end{center}
\end{table*}

\subsection{In-domain Evaluation}
\label{app:static_six}

We report in-domain results on all six benchmarks. Both methods use the same
frozen Chronos-Bolt backbone and the same adapter configuration ($r = 32$,
$\alpha = 128$, all 12 decoder layers, $N_{\mathrm{pg}} = 1$); the only
difference is whether the LoRA matrices are learned once per dataset or
generated per input instance. The static LoRA baseline uses $\eta = 1 \times 10^{-5}$, selected from
\{$1e-5, 1e-6, 1e-7$\} on the validation
split of each target dataset. Both methods train for up to 10 epochs with
early stopping (patience 3) on validation loss.

\begin{table}[t]
\centering
\caption{In-domain comparison of static LoRA and AdaCast across all six
benchmarks ($T{=}512$, $H{=}64$, $r{=}32$, $\alpha{=}128$,
$N_{\mathrm{pg}}{=}1$). Each cell reports MSE\,/\,MAE. \textbf{Bold}
indicates the better result.}
\label{tab:static_six}
\small
\setlength{\tabcolsep}{5pt}
\begin{center}
\begin{tabular}{lcc}
\toprule
Dataset & Chronos-Bolt LoRA & AdaCast \\
\midrule
ETTh1    & 0.3505 / 0.3610 & \textbf{0.3431 / 0.3594} \\
ETTh2    & 0.2412 / 0.2997 & \textbf{0.2370 / 0.2919} \\
ETTm1    & 0.2840 / 0.3155 & \textbf{0.2603 / 0.3040} \\
ETTm2    & 0.1401 / 0.2171 & \textbf{0.1385 / 0.2160} \\
Weather  & 0.1301 / 0.1592 & \textbf{0.1255 / 0.1544} \\
Exchange & 0.0591 / 0.1670 & \textbf{0.0571 / 0.1650} \\
\midrule
Average  & 0.2008 / 0.2533 & \textbf{0.1936 / 0.2485} \\
\bottomrule
\end{tabular}
\end{center}
\end{table}

AdaCast outperforms static adaptation on both metrics across all six
datasets. The margin is largest on ETTm1 ($8.3\%$ MSE) and smallest on
ETTm2 ($1.1\%$). Averaged across benchmarks, instance-conditioned generation
reduces MSE by $3.4\%$ and MAE by $2.0\%$ relative to a single shared
adapter trained on the same target dataset.

\subsection{Evaluation on Longer Forecast Horizons}
\label{app:longer_horizon}
 
\paragraph{Setting.}
All experiments in the main paper use a fixed context length of
$T=512$ and prediction horizon $H=64$, matching the native horizon
Chronos-Bolt's decoder is trained to produce in a single forward pass.
To assess how AdaCast's adaptation behaves at longer horizons, we
additionally evaluate at $T=512$ with $H \in \{64, 96, 192, 336\}$
across three configurations: \textbf{Chronos-Bolt\textsubscript{B} Zero-shot}, \textbf{AdaCast (In-domain)}, and \textbf{AdaCast (Zeroshot)}.
 
\paragraph{Rolling forecast for $H > 64$.}
Since the backbone's decoder is natively trained to emit quantile
predictions only for its trained horizon of 64 steps, horizons beyond
this require an autoregressive extension at inference time. We adopt
the rolling-forecast strategy used by TS-RAG ~\citep{ning2025tsrag}

\subsection{Generator Depth}
We ablate the number of generator blocks, $N_{\mathrm{pg}} \in \{1, 4, 8, 16\}$,
holding $r = 32$, $\alpha = 128$, $\eta = 1e-7$, and the
pretraining corpus fixed. Table~\ref{tab:pgdepth} reports zeroshot results.

Generator depth has little effect on four of the six datasets. On ETTh1,
ETTh2, ETTm1, and ETTm2 the full range across $N_{\mathrm{pg}}$ is at most
$0.0009$ MSE, comparable to the seed-to-seed standard deviations reported in
Table~\ref{tab:seeds}, so these differences are not distinguishable from
run-to-run variation. Exchange is the clear exception: MSE decreases
monotonically from $0.0654$ at $N_{\mathrm{pg}} = 1$ to $0.0591$ at
$N_{\mathrm{pg}} = 16$, a range twenty times the corresponding seed spread,
and Weather MAE follows the same monotone pattern
($0.1891 \rightarrow 0.1785$). In both cases the improvement is largely
exhausted by $N_{\mathrm{pg}} = 8$, with $N_{\mathrm{pg}} = 16$ offering no
further gain at twice the generator parameters. We therefore adopt
$N_{\mathrm{pg}} = 8$ for zeroshot pretraining.

\begin{table}[]
\centering
\caption{
Intervention statistics for decoder blocks 9--11. All quantities except
$\mathrm{+frac}$ are changes in approximate CRPS, in units of $10^{-3}$.
``Rec.'' is the reconstruction-only control; ``Adj.'' is the median minus
reconstruction control. Mean, standard deviation, and maximum are over the 64
ablated features.
}
\label{tab:supp_sae_late_block_stats}
\small
\setlength{\tabcolsep}{3.5pt}
\begin{center}
\resizebox{\columnwidth}{!}{%
\begin{tabular}{llcrrrrrr}
\toprule
& Blk & $\mathrm{+frac}$ & Med. & Mean & Std. & Max. & Rec. & Adj. \\
\midrule
\multirow{3}{*}{ETTh1}
 & 9  & 0.84 & 1.02 & 1.87 & 2.28 & 12.0 & $-0.34$ & 1.36 \\
 & 10 & 1.00 & 10.9 & 10.8 & 5.76 & 41.9 & 1.22 & 9.69 \\
 & 11 & 1.00 & 25.8 & 29.5 & 16.2 & 98.9 & 3.82 & 21.9 \\
\midrule
\multirow{3}{*}{ETTh2}
 & 9  & 0.88 & 1.18 & 1.51 & 2.38 & 18.1 & $-0.08$ & 1.26 \\
 & 10 & 1.00 & 6.87 & 7.60 & 3.86 & 20.8 & 0.13 & 6.75 \\
 & 11 & 1.00 & 20.8 & 41.3 & 140 & 1149 & 0.95 & 19.9 \\
\midrule
\multirow{3}{*}{ETTm1}
 & 9  & 0.94 & 1.35 & 2.66 & 3.19 & 12.5 & 0.06 & 1.29 \\
 & 10 & 1.00 & 7.59 & 11.9 & 16.3 & 101 & 0.11 & 7.48 \\
 & 11 & 1.00 & 21.0 & 22.8 & 10.8 & 53.4 & 0.30 & 20.7 \\
\midrule
\multirow{3}{*}{ETTm2}
 & 9  & 0.92 & 0.71 & 1.64 & 2.20 & 10.6 & 0.00 & 0.71 \\
 & 10 & 1.00 & 5.97 & 8.26 & 9.48 & 60.5 & 0.11 & 5.86 \\
 & 11 & 1.00 & 16.9 & 16.9 & 8.39 & 37.4 & 0.24 & 16.6 \\
\bottomrule
\end{tabular}%
}
\end{center}
\end{table}

\begin{table*}[t]
\centering
\caption{Zeroshot forecasting results for extended forecasting horizons (MSE).}
\small
\label{tab:longer_horizon}
\begin{center}
\begin{tabular}{ll ccccc c}
\toprule
\textbf{Horizon} & \textbf{Methods} & \textbf{ETTh1} & \textbf{ETTh2} & \textbf{ETTm1} & \textbf{ETTm2} & \textbf{Weather} & \textbf{Exchange} \\
\midrule
\multirow{3}{*}{64}
& Chronos Zeroshot    & 0.3616 & 0.2517 & 0.3109 & 0.1487 & 0.1525 & 0.0673 \\
& AdaCast (In-domain)   & \textbf{0.3431} & \textbf{0.2370} & \textbf{0.2603} & \textbf{0.1385} & \textbf{0.1255} & \textbf{0.0571} \\
& AdaCast (Zero-shot)   & 0.3593 & 0.2450 & 0.3017 & 0.1485 & 0.1491 & 0.0594 \\
\midrule
\multirow{3}{*}{96}
& Chronos Zeroshot    & 0.3859 & 0.2899 & 0.3323 & 0.1779 & 0.1777 & 0.0993 \\
& AdaCast (In-domain)   & \textbf{0.3691} & \textbf{0.2739} & \textbf{0.2891} & \textbf{0.1672} & \textbf{0.1459} & \textbf{0.0870} \\
& AdaCast (Zero-shot)   & 0.3798 & 0.2789 & 0.3259 & 0.1781 & 0.1763 & 0.0918 \\
\midrule
\multirow{3}{*}{192}
& Chronos Zeroshot    & 0.4446 & 0.3603 & 0.3838 & 0.2515 & 0.2244 & 0.1926 \\
& AdaCast (In-domain)   & \textbf{0.4233} & \textbf{0.3462} & \textbf{0.3447} & \textbf{0.2336} & \textbf{0.1932} & \textbf{0.1738} \\
& AdaCast (Zero-shot)   & 0.4380 & 0.3547 & 0.3853 & 0.2486 & 0.2310 & 0.1850 \\
\midrule
\multirow{3}{*}{336}
& Chronos Zeroshot    & 0.4850 & 0.4045 & 0.4374 & 0.3177 & 0.2838 & 0.3437 \\
& AdaCast (In-domain)   & \textbf{0.4437} & \textbf{0.3740} & \textbf{0.3859} & \textbf{0.2893} & \textbf{0.2452} & \textbf{0.3059} \\
& AdaCast (Zero-shot)   & 0.4771 & 0.4029 & 0.4249 & 0.3203 & 0.2733 & 0.3339 \\
\bottomrule
\end{tabular}
\end{center}
\end{table*}

\section{Sparse-Feature Sensitivity and Decoder-Region Adaptation}
\label{sec:supp_sae}

We examine whether decoder blocks that are sensitive to representation
perturbations are also effective locations for AdaCast's generated
updates. For each ETT dataset and each of the 12 frozen Chronos-Bolt
decoder blocks, we train a separate TopK sparse autoencoder
(SAE)~\cite{Mishra2026DissectingChronos,Gao2024ScalingSAE,
Templeton2024ScalingMonosemanticity} using activations collected only
from the corresponding training split. We then ablate individual SAE
features, measure the resulting change in forecast quality, and compare
the resulting sensitivity profile with the performance of AdaCast when
generated updates are restricted to different decoder regions.
Section~\ref{sec:supp_sae_protocol} describes the SAE protocol,
Sections~\ref{sec:supp_sae_recon}--\ref{sec:supp_sae_magnitude} report
reconstruction quality and intervention effects, and
Section~\ref{sec:supp_sae_region} relates these effects to
decoder-region adaptation.

%--------------------------------------------------------------------
\subsection{SAE Training Protocol}
\label{sec:supp_sae_protocol}

\paragraph{Activation collection.}
For each of the four ETT datasets, we use only the standard
chronological training split and construct context--target windows
with context length $T=512$ and prediction horizon $H=64$. Each
variable presented to Chronos-Bolt is treated as an independent
univariate series. The Chronos-Bolt Base backbone is placed in
evaluation mode and remains frozen throughout the experiment. For
each of its 12 decoder blocks, we collect the first tensor returned by
the block, corresponding to its output hidden state
$\mathbf{h}\in\mathbb{R}^{d}$. Activations are flattened over
forecasting examples and decoder positions, and a separate SAE is
trained for every dataset--block pair.

\paragraph{TopK SAE parameterization.}
Before encoding, each block's activations are rescaled using
\begin{equation}
s
=
\frac{\sqrt{d}}
{\mathbb{E}_{i}\!\left[\lVert\mathbf{h}_{i}\rVert_{2}\right]},
\qquad
\widetilde{\mathbf{h}}=s\mathbf{h}.
\label{eq:supp_sae_scale}
\end{equation}
The encoder is
\begin{equation}
\mathbf{z}
=
\operatorname{TopK}_{K}
\left(
\operatorname{ReLU}
\left(
W_{\mathrm{enc}}
(\widetilde{\mathbf{h}}-\mathbf{b}_{\mathrm{pre}})
+\mathbf{b}_{\mathrm{lat}}
\right)
\right),
\label{eq:supp_sae_encoder}
\end{equation}
and the reconstruction in the scaled activation space is
\begin{equation}
\widehat{\widetilde{\mathbf{h}}}
=
W_{\mathrm{dec}}\mathbf{z}
+\mathbf{b}_{\mathrm{pre}},
\qquad
\widehat{\mathbf{h}}
=
\frac{\widehat{\widetilde{\mathbf{h}}}}{s}.
\label{eq:supp_sae_decoder}
\end{equation}
The latent width is $d_{\mathrm{SAE}}=3072$, and the largest
$K=64$ nonnegative latent coordinates are retained for each hidden
state. The decoder is initialized from the transpose of the encoder,
after which the two matrices are optimized independently. Decoder
columns are constrained to unit norm by projecting out their
parallel-gradient components and renormalizing them after every
optimization step. The full configuration is listed in
Table~\ref{tab:supp_sae_configuration}.

\begin{table}[t]
\centering
\caption{
SAE configuration used for every dataset--block pair. The backbone is
frozen, and only the SAE parameters are optimized.
}
\label{tab:supp_sae_configuration}
\small
\begin{center}
\begin{tabular}{l l}
\toprule
Setting & Value \\
\midrule
Backbone & Chronos-Bolt Base \\
Decoder blocks & $0,\ldots,11$ \\
Latent width $d_{\mathrm{SAE}}$ & $3072$ \\
TopK sparsity $K$ & $64$ \\
Auxiliary TopK width & $96$ \\
Optimizer & Adam, $\beta_1=0.9$, $\beta_2=0.999$ \\
Learning rate & $3\times10^{-4}$ \\
Schedule & Cosine annealing \\
Training epochs & $5$ \\
Batch size & $32$ activation vectors \\
Auxiliary-loss weight & $1/32$ \\
Dead-feature threshold & $10\%$ of optimization steps \\
Gradient clipping & $1.0$ \\
Feature interventions & Top $64$ by mean activation \\
\bottomrule
\end{tabular}
\end{center}
\end{table}

\paragraph{SAE optimization.}
The primary SAE objective is the mean-squared reconstruction error in
the scaled activation space,
\begin{equation}
\mathcal{L}_{\mathrm{recon}}
=
\frac{1}{Nd}
\sum_{i=1}^{N}
\left\|
\widetilde{\mathbf{h}}_{i}
-
\widehat{\widetilde{\mathbf{h}}}_{i}
\right\|_{2}^{2}.
\label{eq:supp_sae_recon_loss}
\end{equation}
To reduce dead latents, a feature is marked dead after remaining
inactive for $10\%$ of the optimization run. When dead features are
present, up to 96 dead latents are additionally trained to reconstruct
the detached residual
$\widetilde{\mathbf{h}}-\widehat{\widetilde{\mathbf{h}}}$ with
weight $1/32$. The total loss is therefore
\begin{equation}
\mathcal{L}_{\mathrm{SAE}}
=
\mathcal{L}_{\mathrm{recon}}
+
\frac{1}{32}\mathcal{L}_{\mathrm{aux}},
\label{eq:supp_sae_total_loss}
\end{equation}
where $\mathcal{L}_{\mathrm{aux}}=0$ when no latent satisfies the
dead-feature criterion.

%--------------------------------------------------------------------
\subsection{Reconstruction Quality}
\label{sec:supp_sae_recon}

Reconstruction quality is measured on up to $4096$ randomly sampled
training activation vectors using the fraction of variance
unexplained,
\begin{equation}
\mathrm{FVU}
=
\frac{
    \sum_i
    \left\|
    \widetilde{\mathbf{h}}_i-
    \widehat{\widetilde{\mathbf{h}}}_i
    \right\|_2^2
}{
    \sum_i
    \left\|
    \widetilde{\mathbf{h}}_i-
    \overline{\widetilde{\mathbf{h}}}
    \right\|_2^2
},
\qquad
\mathrm{FVE}=1-\mathrm{FVU},
\label{eq:supp_fve}
\end{equation}
where $\overline{\widetilde{\mathbf{h}}}
=\frac{1}{N}\sum_{i=1}^{N}\widetilde{\mathbf{h}}_i$.
Because the scaling factor $s$ is shared within a dataset--block
pair, it cancels from the FVU ratio.

\begin{figure*}[t]
    \centering
    \includegraphics[width=0.9\linewidth]
    {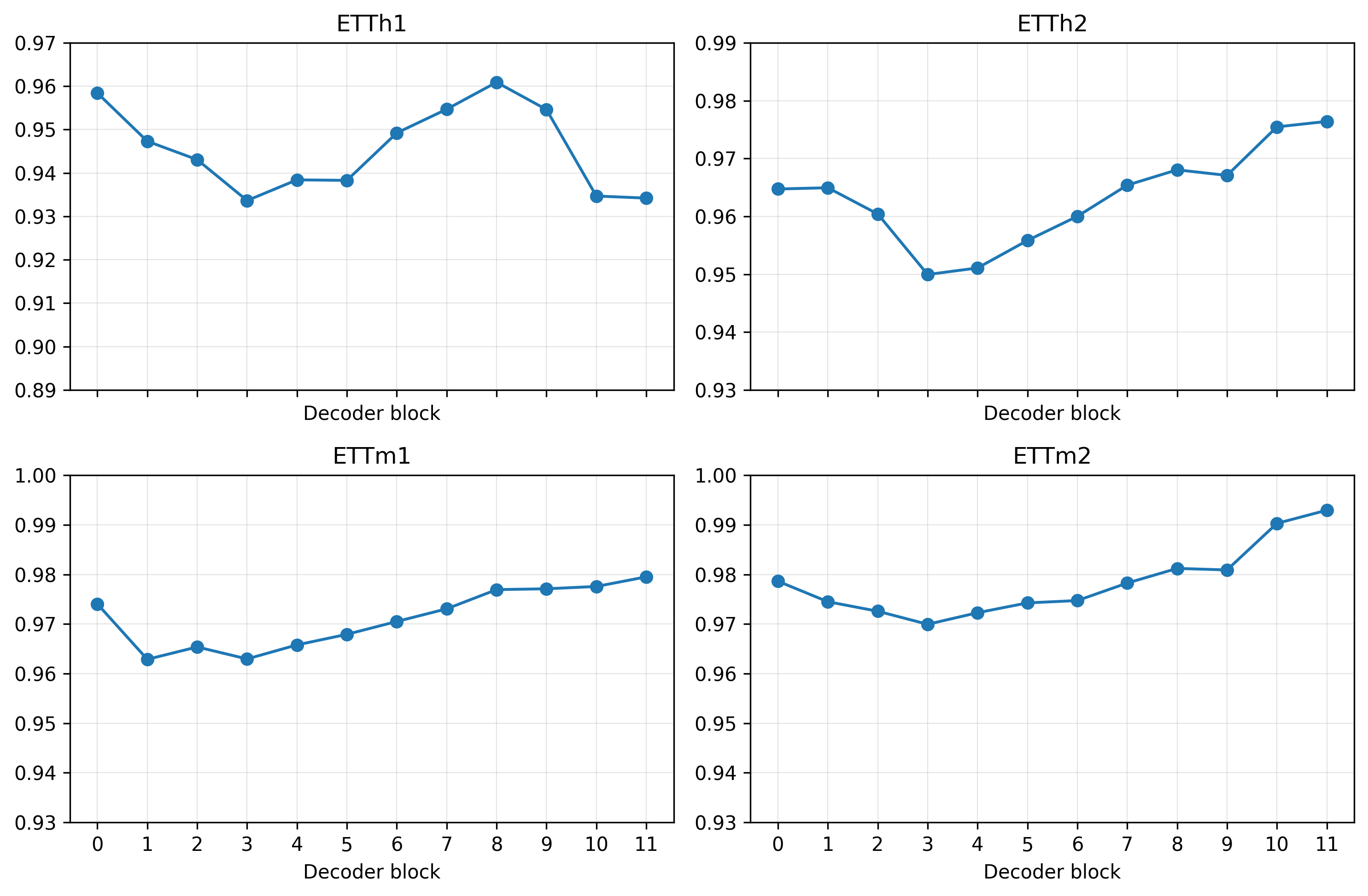}
    \caption{
    Fraction of variance explained
    ($\mathrm{FVE}=1-\mathrm{FVU}$) by the TopK SAE trained at each
    Chronos-Bolt decoder block. A separate SAE is trained for every
    dataset--block pair using activations from the corresponding
    chronological training split.
    }
    \label{fig:supp_sae_fve_all}
\end{figure*}

Figure~\ref{fig:supp_sae_fve_all} reports FVE for every
dataset--block pair. On ETTh1, FVE ranges from $0.934$ to $0.961$
across decoder blocks, and on ETTm1 from $0.963$ to $0.980$. The SAEs
therefore retain most of the variance in the original block
activations, so feature-level interventions operate on a
representation that closely approximates the original hidden state.

%--------------------------------------------------------------------
\subsection{Feature Interventions and Sign Statistic}
\label{sec:supp_sae_sign}

\paragraph{Feature ranking and ablation.}
For each trained SAE, features are ranked by their mean latent
activation over all collected training activations,
\begin{equation}
a_j
=
\frac{1}{N}
\sum_{i=1}^{N} z_{ij},
\label{eq:supp_sae_feature_rank}
\end{equation}
where inactive entries are included as zeros. We retain the 64
highest-ranked features and ablate them individually. For feature
$j$, the intervention is
\begin{equation}
z_j\leftarrow 0.
\label{eq:supp_sae_ablation}
\end{equation}
The intervened latent vector is decoded through
Equation~\ref{eq:supp_sae_decoder}, and the reconstructed hidden state
is patched into the corresponding frozen decoder block. Both feature
ranking and intervention evaluation use the chronological training
split; the experiment therefore measures within-distribution
perturbation sensitivity rather than held-out SAE generalization.

\paragraph{Forecast-quality change.}
For every feature intervention, we compute
\begin{equation}
\Delta\widehat{\mathrm{CRPS}}_j
=
\widehat{\mathrm{CRPS}}_{\mathrm{patch}(j)}
-
\widehat{\mathrm{CRPS}}_{\mathrm{original}},
\label{eq:supp_delta_crps}
\end{equation}
where the nine quantiles produced by Chronos-Bolt are evaluated using
\begin{equation}
\widehat{\mathrm{CRPS}}
=
\frac{2}{9}
\sum_{\tau\in\{0.1,\ldots,0.9\}}
\mathrm{QL}_{\tau}
(\widehat{y}_{\tau},y).
\label{eq:supp_crps}
\end{equation}
The quantile loss is averaged over all evaluated windows and forecast
steps. Equation~\ref{eq:supp_crps} is a finite-quantile approximation
and does not characterize the predictive distribution outside the
$0.1$ and $0.9$ quantile boundaries.

\begin{figure}[t]
    \centering
    \includegraphics[width=\linewidth]
    {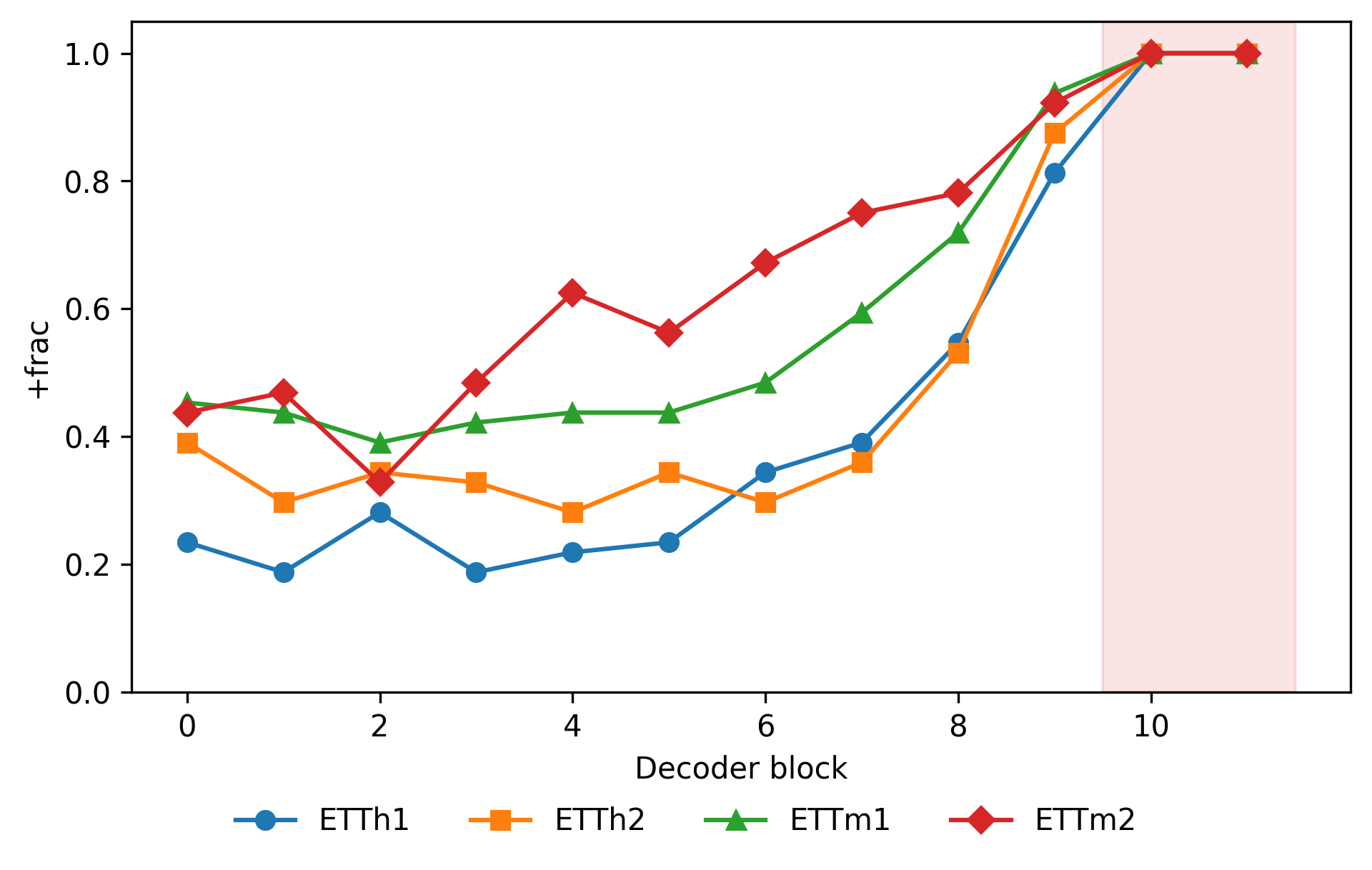} % TODO: set to the former main-paper +frac panel
    \caption{
    Fraction of the Top-64 SAE-feature ablations that increase
    approximate CRPS ($\mathrm{+frac}$,
    Equation~\ref{eq:supp_plus_frac}) at each Chronos-Bolt decoder
    block. All four datasets reach $\mathrm{+frac}=1.00$ at blocks
    10--11.
    }
    \label{fig:supp_sae_plus_frac}
\end{figure}

\paragraph{Sign statistic.}
We summarize the 64 interventions at each block by
\begin{equation}
\mathrm{+frac}
=
\frac{1}{64}
\sum_{j=1}^{64}
\mathbb{I}
\left[
\Delta\widehat{\mathrm{CRPS}}_j>0
\right],
\label{eq:supp_plus_frac}
\end{equation}
the fraction of feature removals that increase approximate CRPS.
Across all four ETT datasets, $\mathrm{+frac}=1.00$ at blocks 10 and
11, meaning that every one of the 64 tested high-activation feature
removals degrades the forecast. At block 9, $\mathrm{+frac}$ ranges
from $0.844$ to $0.938$, while earlier blocks exhibit lower and more
variable values (Figure~\ref{fig:supp_sae_plus_frac}).

\begin{figure*}[t]
\centering
\begin{minipage}[t]{0.45\textwidth}
    \centering
    \includegraphics[width=\linewidth]
    {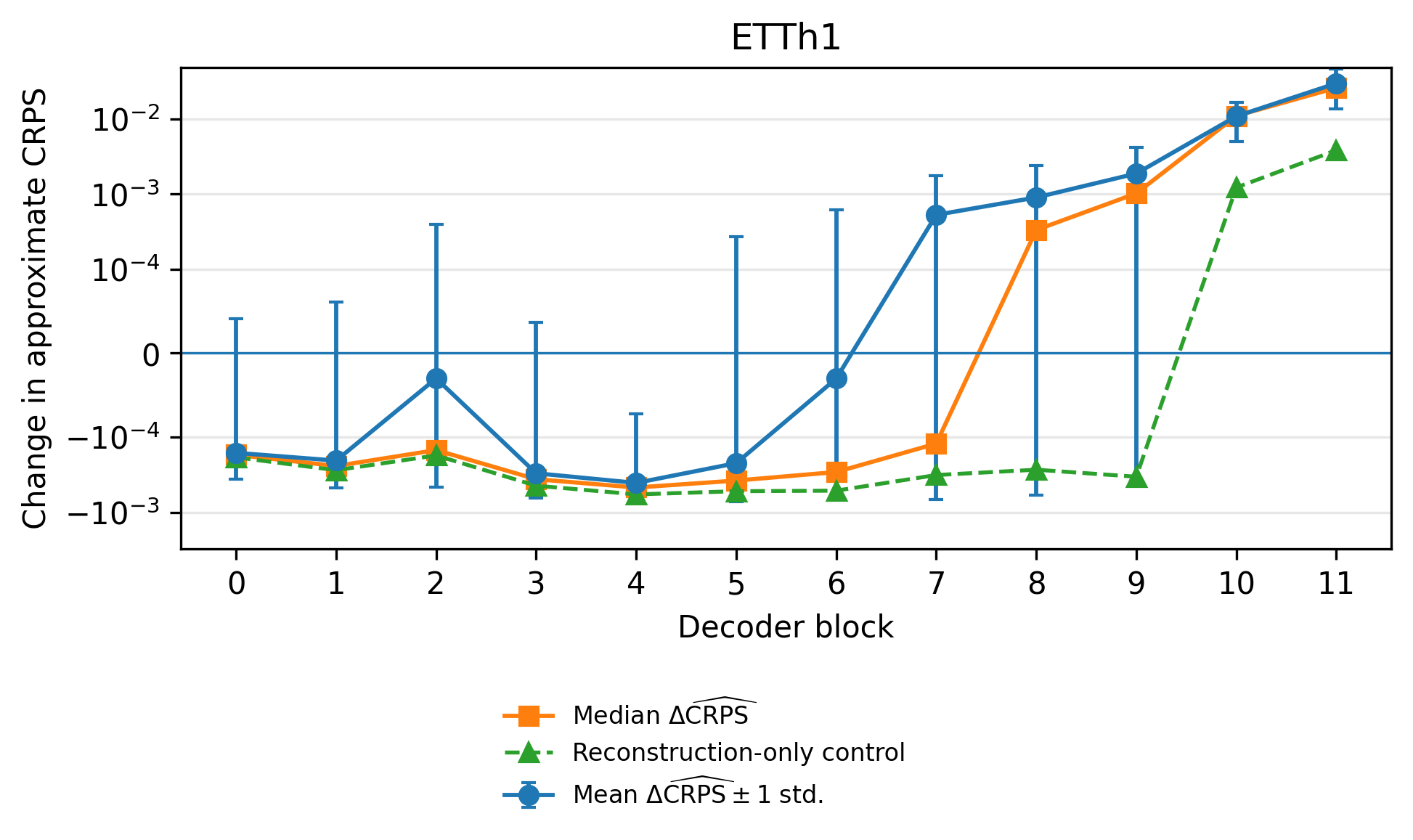}
\end{minipage}
\hfill
\begin{minipage}[t]{0.45\textwidth}
    \centering
    \includegraphics[width=\linewidth]
    {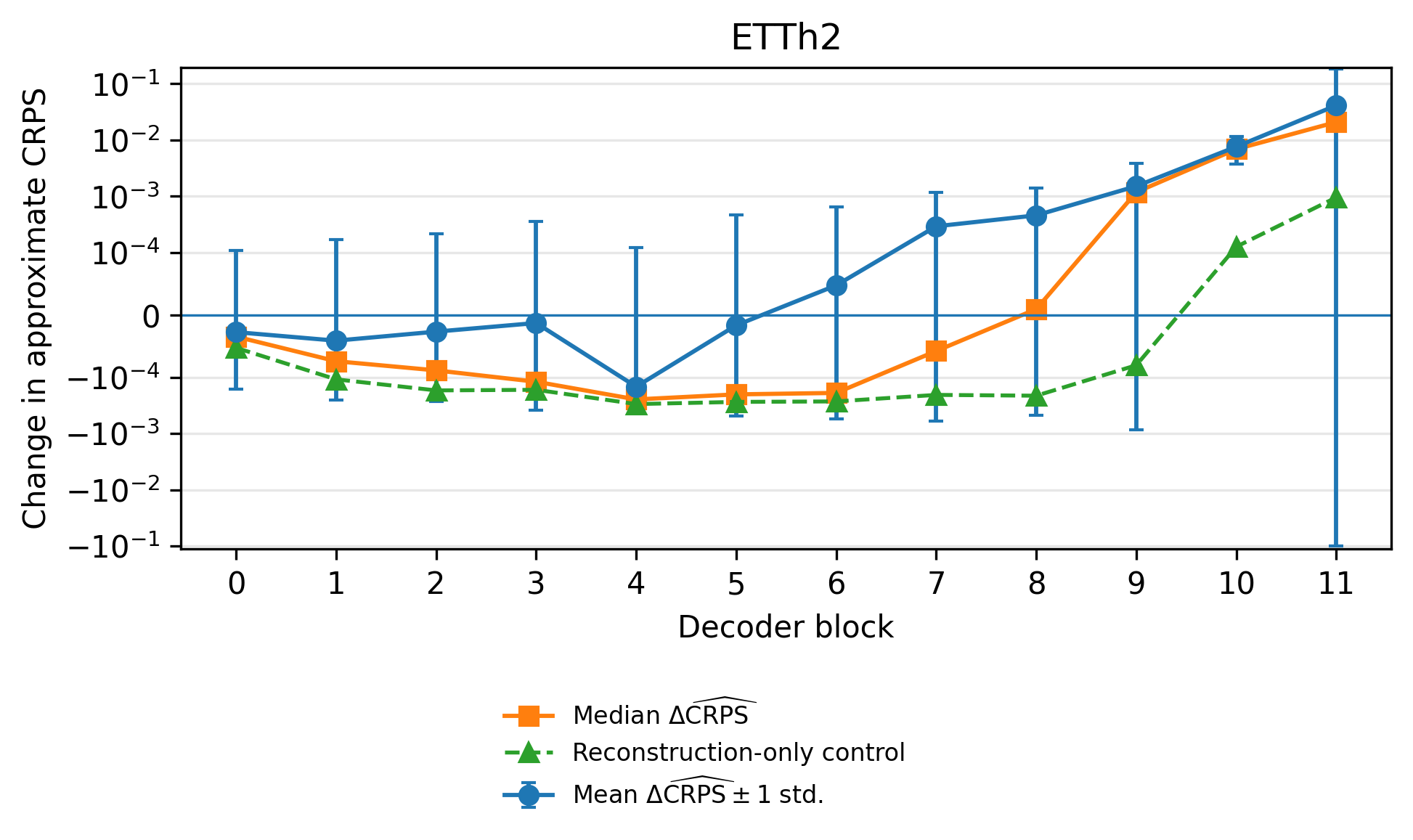}
\end{minipage}

\vspace{0.5em}

\begin{minipage}[t]{0.45\textwidth}
    \centering
    \includegraphics[width=\linewidth]
    {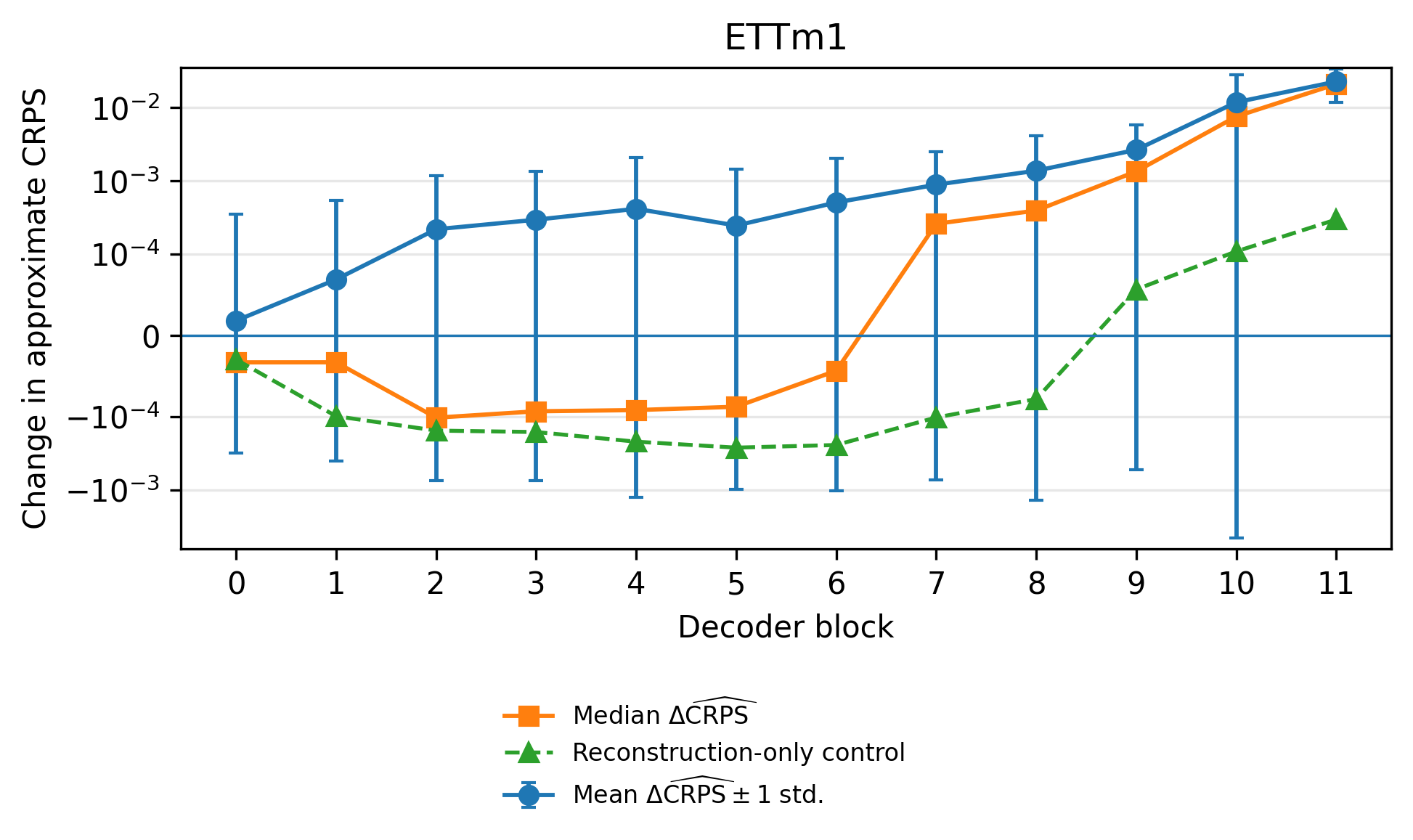}
\end{minipage}
\hfill
\begin{minipage}[t]{0.45\textwidth}
    \centering
    \includegraphics[width=\linewidth]
    {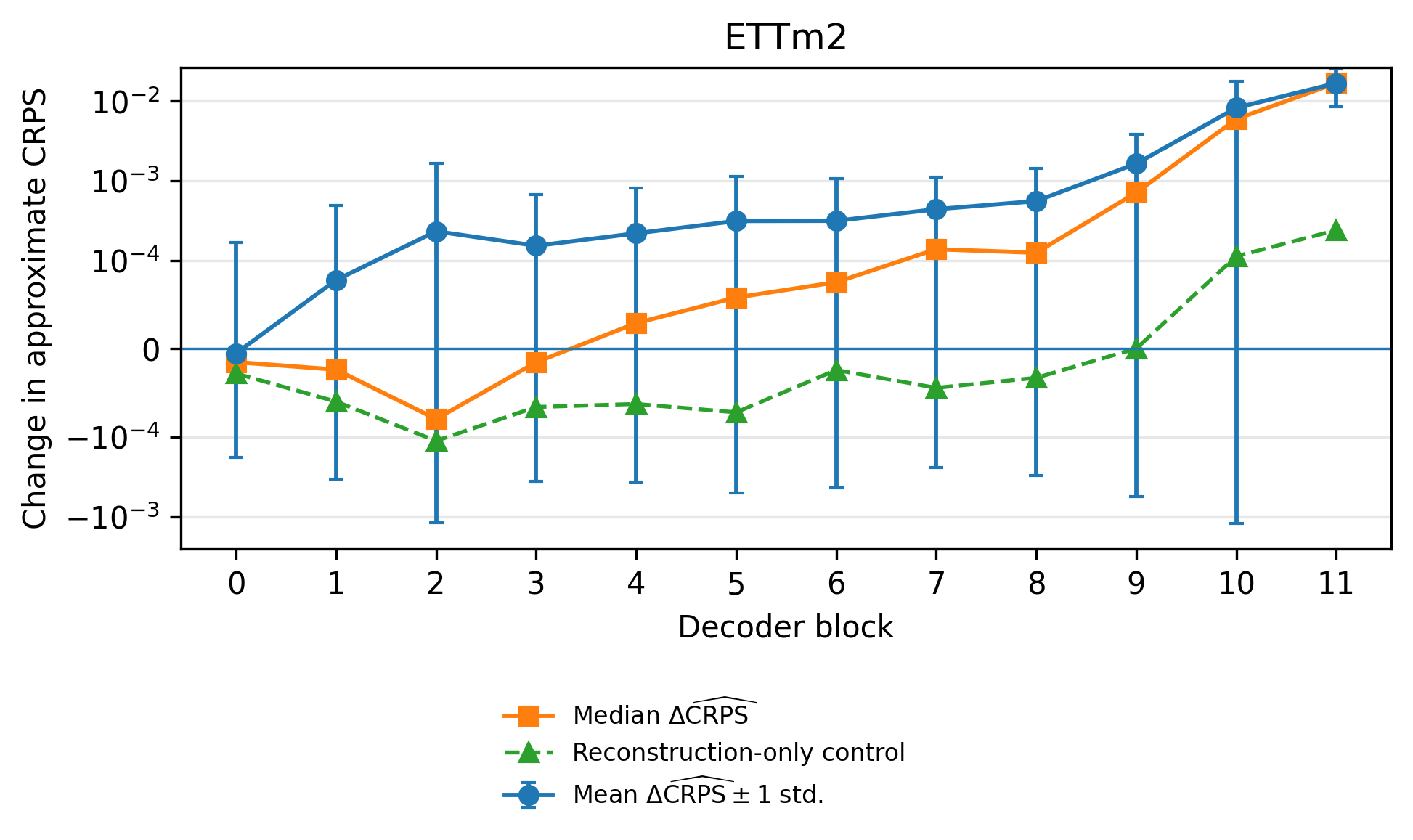}
\end{minipage}
\caption{
Magnitude of the Top-64 SAE-feature interventions across decoder
depth. Points show the mean and median
$\Delta\widehat{\mathrm{CRPS}}$ values; error bars show one standard
deviation across the 64 individually ablated features. The dashed
curve is the reconstruction-only control from
Equation~\ref{eq:supp_delta_crps_recon}.
}
\label{fig:supp_sae_intervention_magnitude}
\end{figure*}

\begin{figure}[t]
    \centering
    \includegraphics[width=\linewidth]
    {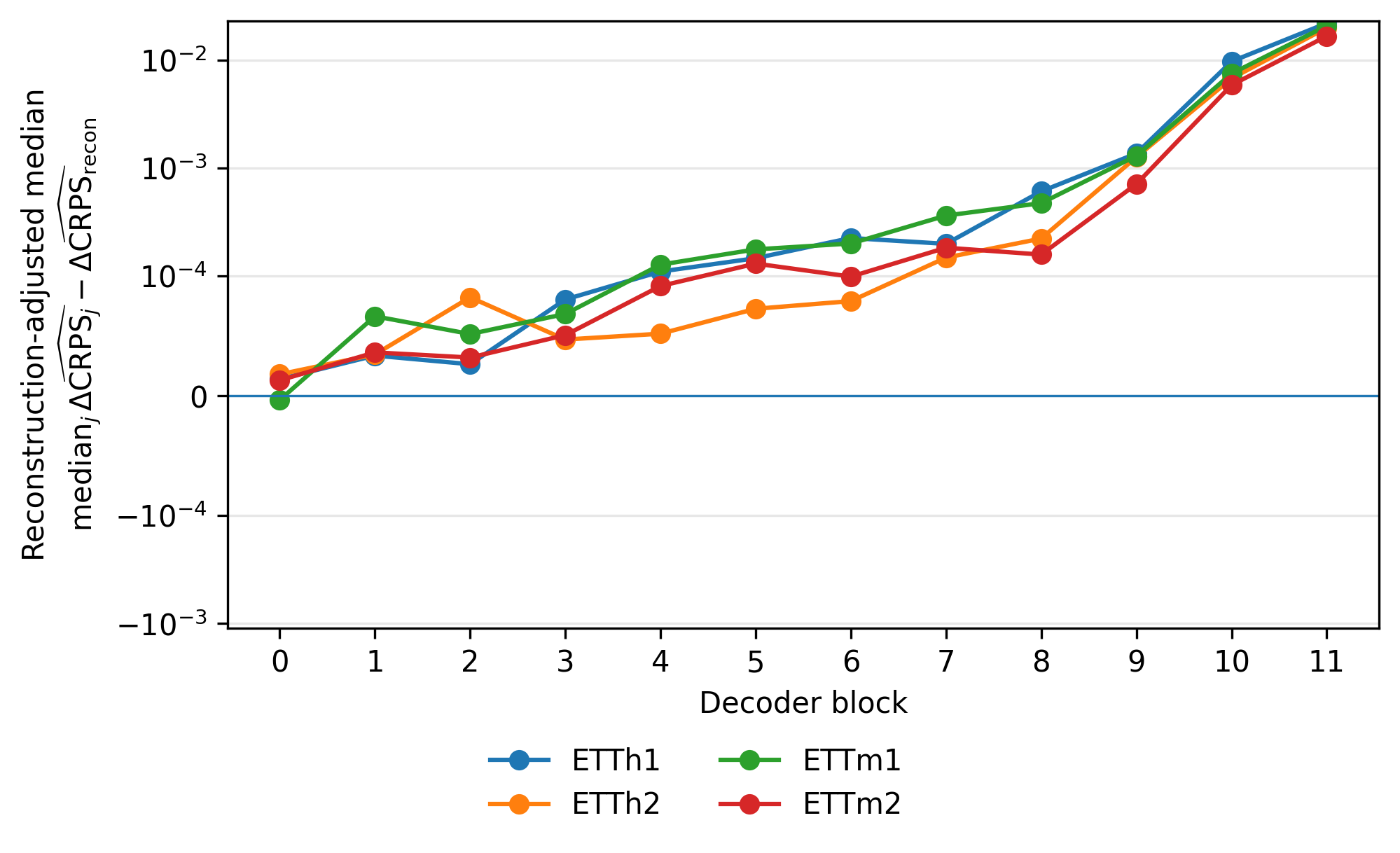}
    \caption{
    Reconstruction-adjusted median intervention effect from
    Equation~\ref{eq:supp_delta_crps_adjusted}. Values remain close to
    zero in early blocks and increase sharply in blocks 9--11 across
    all four datasets.
    }
    \label{fig:supp_sae_adjusted_median}
\end{figure}
%--------------------------------------------------------------------
\subsection{Intervention Magnitude and Reconstruction-Only Control}
\label{sec:supp_sae_magnitude}

Because $\mathrm{+frac}$ discards effect magnitude, we additionally
report the mean, median, standard deviation, and maximum of the 64
$\Delta\widehat{\mathrm{CRPS}}_j$ values for every block.

\paragraph{Reconstruction-only control.}
Encoding and decoding a hidden state can alter the forecast even when
no feature is removed. We therefore perform a reconstruction-only
intervention in which the TopK latent representation is decoded and
patched without zeroing any latent coordinate:
\begin{equation}
\Delta\widehat{\mathrm{CRPS}}_{\mathrm{recon}}
=
\widehat{\mathrm{CRPS}}_{\mathrm{reconstructed}}
-
\widehat{\mathrm{CRPS}}_{\mathrm{original}}.
\label{eq:supp_delta_crps_recon}
\end{equation}
This value is computed independently for every dataset--block pair.
For a direct magnitude comparison, we also report the
reconstruction-adjusted median
\begin{equation}
\Delta\widehat{\mathrm{CRPS}}_{\mathrm{adj}}
=
\operatorname{median}_{j}
\left(
\Delta\widehat{\mathrm{CRPS}}_j
\right)
-
\Delta\widehat{\mathrm{CRPS}}_{\mathrm{recon}}.
\label{eq:supp_delta_crps_adjusted}
\end{equation}
This difference is a descriptive control-adjusted statistic, not an
additive decomposition of the intervention effect. We prefer it to a
ratio because ratios become unstable when the reconstruction-only
control is close to zero.

\paragraph{Results.}
The magnitude analysis is consistent with the sign pattern in
Figure~\ref{fig:supp_sae_plus_frac}
(Figures~\ref{fig:supp_sae_intervention_magnitude}
and~\ref{fig:supp_sae_adjusted_median}). Across datasets, the raw
median intervention effect is small and of mixed sign through most of
blocks 0--8, becomes consistently positive at block 9, and increases
sharply at blocks 10--11.

The reconstruction-only controls are substantially smaller than the
late-block intervention medians. After subtracting the corresponding
control, the median effect ranges from $0.00586$ to $0.00969$ at block
10 and from $0.01664$ to $0.02193$ at block 11. Equivalently, the
absolute median intervention at blocks 10--11 is between $6.7$ and
$70.8$ times the absolute reconstruction-only change. The late-block
increase therefore cannot be explained by SAE reconstruction error
alone. Together, the sign and magnitude analyses identify blocks
10--11 as the most perturbation-sensitive region under the tested SAE
intervention.

%--------------------------------------------------------------------
\subsection{Perturbation Sensitivity versus Adaptation Utility}
\label{sec:supp_sae_region}

We next ask whether the most perturbation-sensitive region is also the
most useful location for AdaCast's generated LoRA updates.
Table~\ref{tab:sae_region} compares full AdaCast, the best tested
contiguous restricted region, and a restriction to blocks 9--11, which
covers the sensitivity peak. On all four datasets, restricting the
parameter generator to blocks 9--11 still improves over zero-shot
Chronos-Bolt but is weaker than the best restricted region in both MSE
and MAE. This reveals a consistent mismatch between
\emph{perturbation sensitivity} and \emph{adaptation utility}: the
representations whose disruption most frequently degrades the frozen
model are not necessarily the most effective locations for generated
parameter updates.

\begin{table}[t]
\centering
\caption{
Decoder-region adaptation on the ETT datasets. ``Best restricted''
denotes the best tested contiguous region, shown in brackets.
Blocks 9--11 cover the sensitivity peak identified by the SAE
diagnostic. Each entry reports MSE/MAE.
}
\label{tab:sae_region}
\small
\begin{center}
\resizebox{\columnwidth}{!}{%
\begin{tabular}{l c c c c}
\toprule
Dataset &
\shortstack{Chronos-Bolt\\Zero-shot} &
\shortstack{Full\\AdaCast} &
\shortstack{Best\\restricted} &
\shortstack{Blocks\\9--11} \\
\midrule
ETTh1 &
0.3616/0.3650 &
0.3431/0.3594 &
0.3453/0.3584~[0--3] &
0.3506/0.3620 \\
ETTh2 &
0.2517/0.2992 &
0.2370/0.2919 &
0.2364/0.2919~[6--9] &
0.2394/0.2949 \\
ETTm1 &
0.3109/0.3185 &
0.2603/0.3040 &
0.2621/0.3043~[0--8] &
0.2798/0.3084 \\
ETTm2 &
0.1487/0.2236 &
0.1385/0.2160 &
0.1375/0.2156~[0--3] &
0.1409/0.2182 \\
\bottomrule
\end{tabular}%
}
\end{center}

\end{table}

One plausible explanation is that the final blocks form a
task-specialized region close to the forecasting head. Updates
introduced there directly affect the predicted quantiles and undergo
little subsequent processing by frozen decoder layers. Updates applied
to earlier or middle blocks, in contrast, can be transformed by
downstream frozen blocks before reaching the forecasting head. High
SAE-based sensitivity may therefore indicate limited tolerance to
imperfect updates.
\section{Additional Qualitative Forecasting Results}
\label{sec:additional_qualitative_results}

\vspace{-0.5em}

\noindent
\begin{minipage}{\columnwidth}
    \centering
    \includegraphics[
        width=\linewidth,
        height=0.19\textheight,
        keepaspectratio
    ]{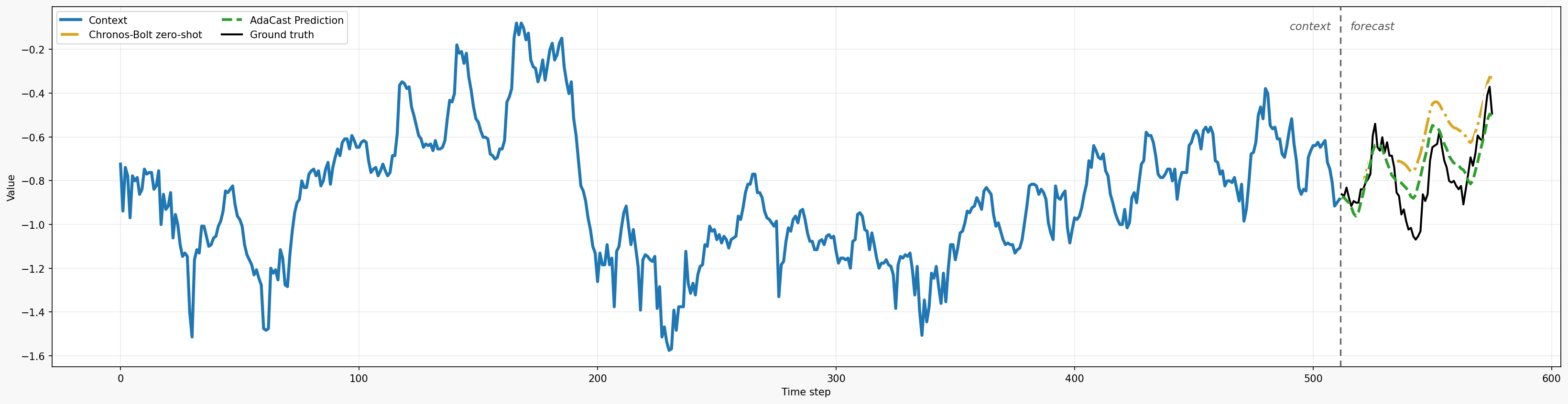}

    \vspace{-0.4em}
    {\small ETTh1, Sample 1}
\end{minipage}

\vspace{0.5em}

\noindent
\begin{minipage}{\columnwidth}
    \centering
    \includegraphics[
        width=\linewidth,
        height=0.19\textheight,
        keepaspectratio
    ]{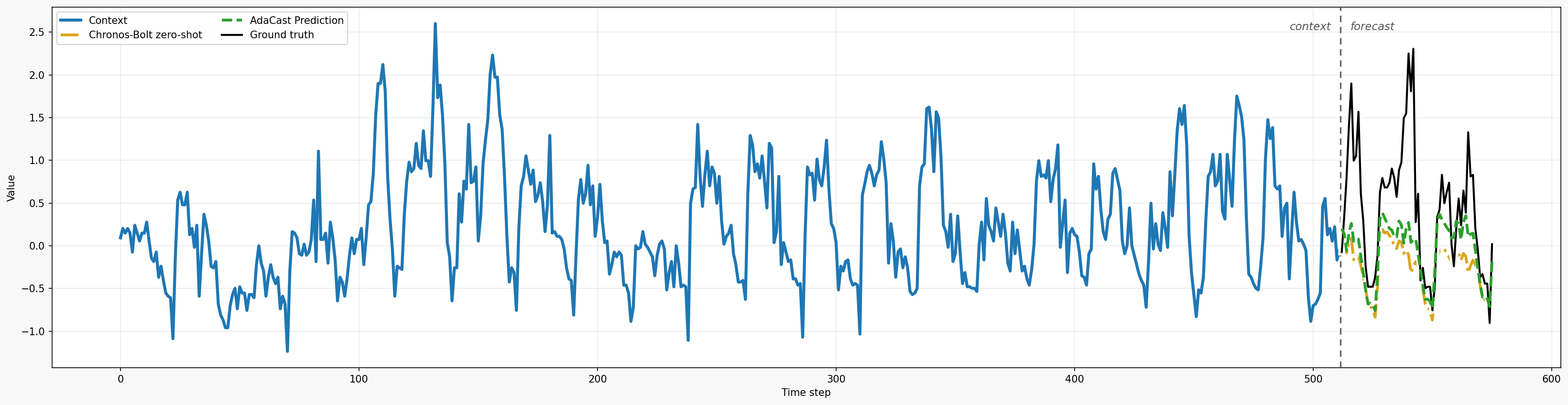}

    \vspace{-0.4em}
    {\small ETTh1, Sample 2}
\end{minipage}

\vspace{0.5em}

\noindent
\begin{minipage}{\columnwidth}
    \centering
    \includegraphics[
        width=\linewidth,
        height=0.19\textheight,
        keepaspectratio
    ]{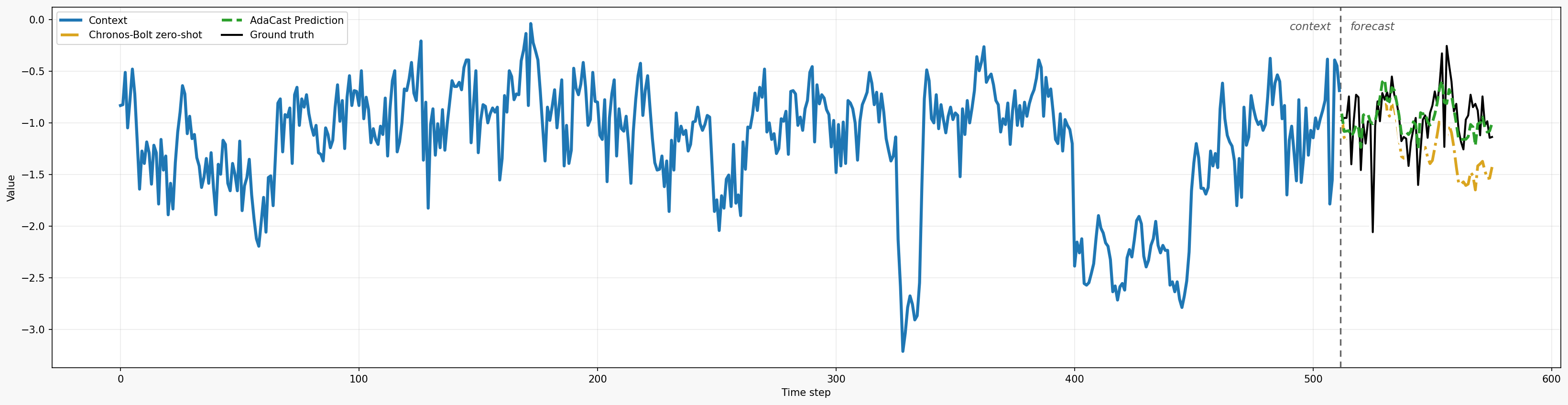}

    \vspace{-0.4em}
    {\small ETTh2, Sample 1}
\end{minipage}

\vspace{0.5em}

\noindent
\begin{minipage}{\columnwidth}
    \centering
    \includegraphics[
        width=\linewidth,
        height=0.19\textheight,
        keepaspectratio
    ]{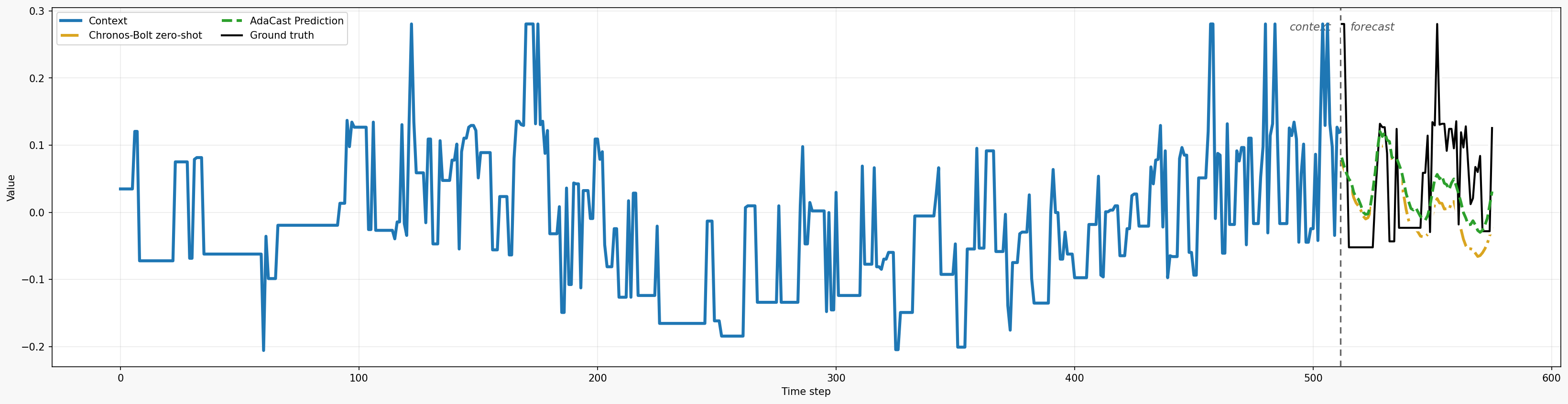}

    \vspace{-0.4em}
    {\small ETTh2, Sample 2}
\end{minipage}

\vspace{0.5em}

\noindent
\begin{minipage}{\columnwidth}
    \centering
    \includegraphics[
        width=\linewidth,
        height=0.19\textheight,
        keepaspectratio
    ]{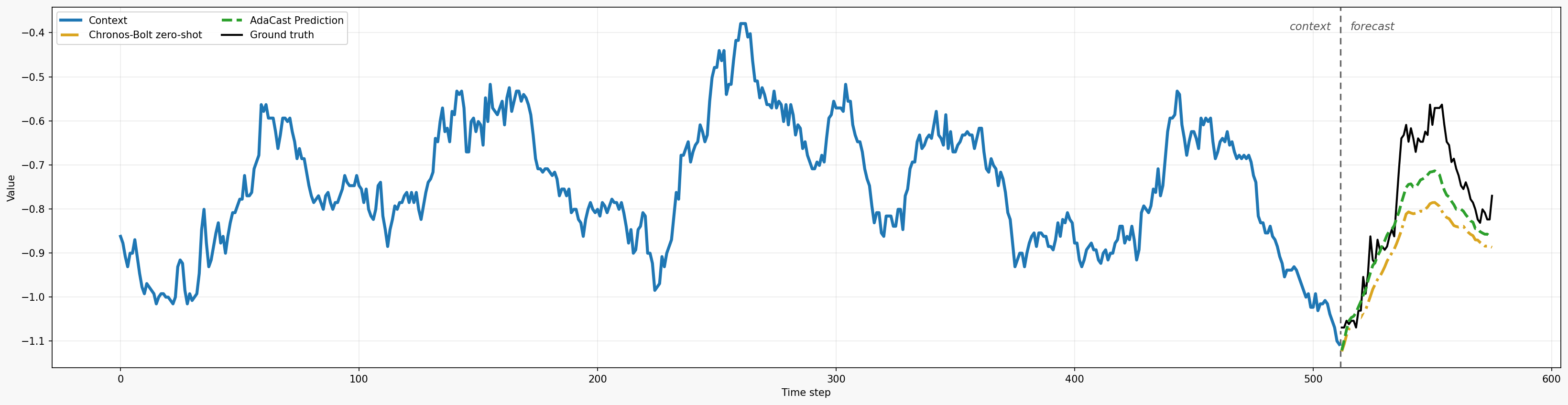}

    \vspace{-0.4em}
    {\small ETTm1, Sample 1}
\end{minipage}

\vspace{0.5em}

\noindent
\begin{minipage}{\columnwidth}
    \centering
    \includegraphics[
        width=\linewidth,
        height=0.19\textheight,
        keepaspectratio
    ]{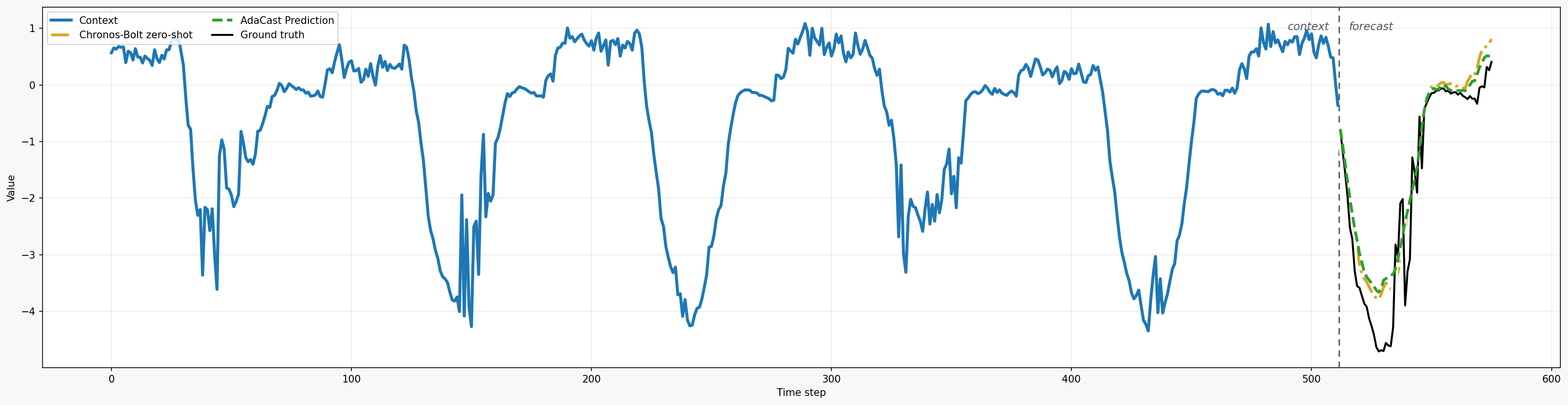}

    \vspace{-0.4em}
    {\small ETTm1, Sample 2}
\end{minipage}

\vspace{0.5em}

\noindent
\begin{minipage}{\columnwidth}
    \centering
    \includegraphics[
        width=\linewidth,
        height=0.19\textheight,
        keepaspectratio
    ]{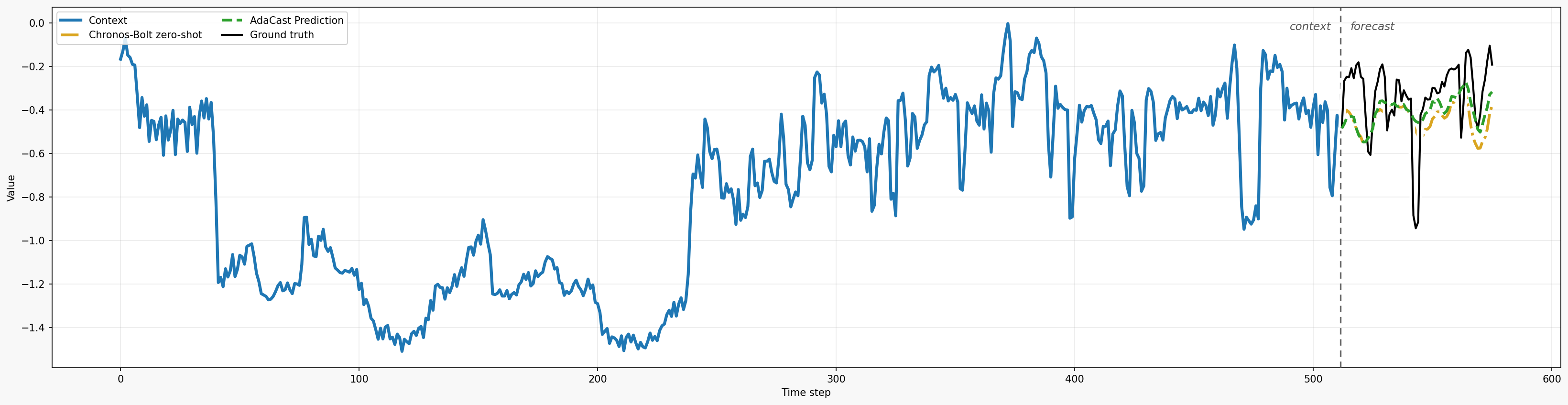}

    \vspace{-0.4em}
    {\small ETTm2, Sample 1}
\end{minipage}

\vspace{0.5em}

\noindent
\begin{minipage}{\columnwidth}
    \centering
    \includegraphics[
        width=\linewidth,
        height=0.19\textheight,
        keepaspectratio
    ]{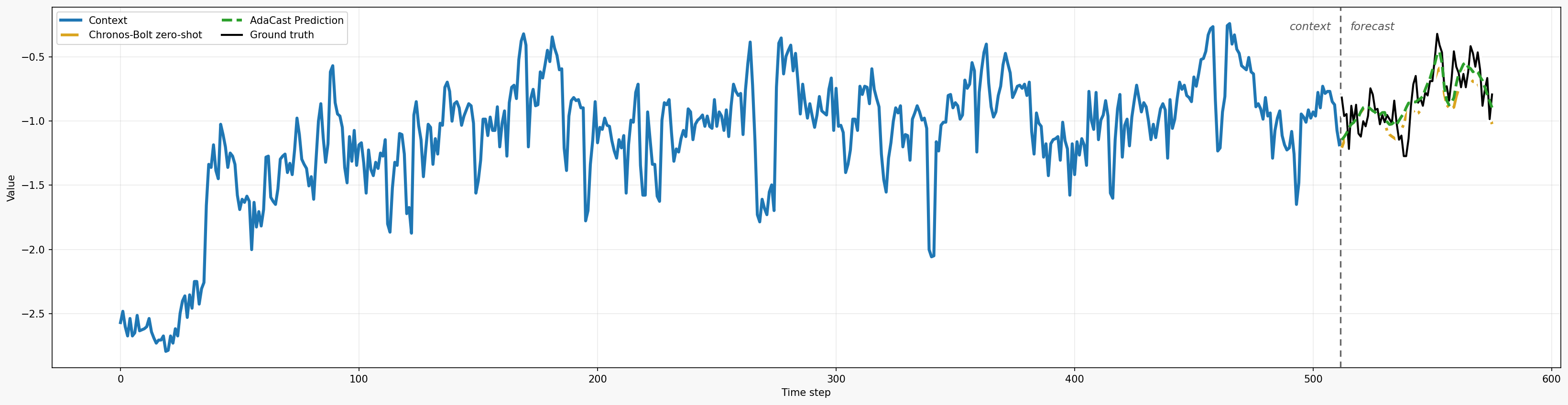}

    \vspace{-0.4em}
    {\small ETTm2, Sample 2}
\end{minipage}

\vspace{0.5em}

\noindent
\begin{minipage}{\columnwidth}
    \centering
    \includegraphics[
        width=\linewidth,
        height=0.19\textheight,
        keepaspectratio
    ]{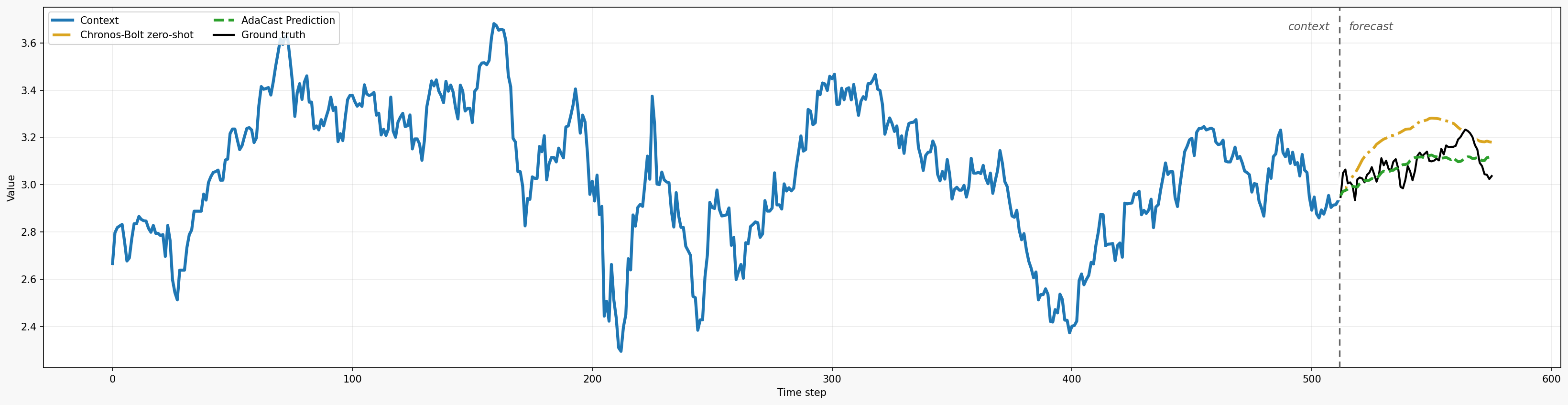}

    \vspace{-0.4em}
    {\small Exchange, Sample 1}
\end{minipage}

\vspace{0.5em}

\noindent
\begin{minipage}{\columnwidth}
    \centering
    \includegraphics[
        width=\linewidth,
        height=0.19\textheight,
        keepaspectratio
    ]{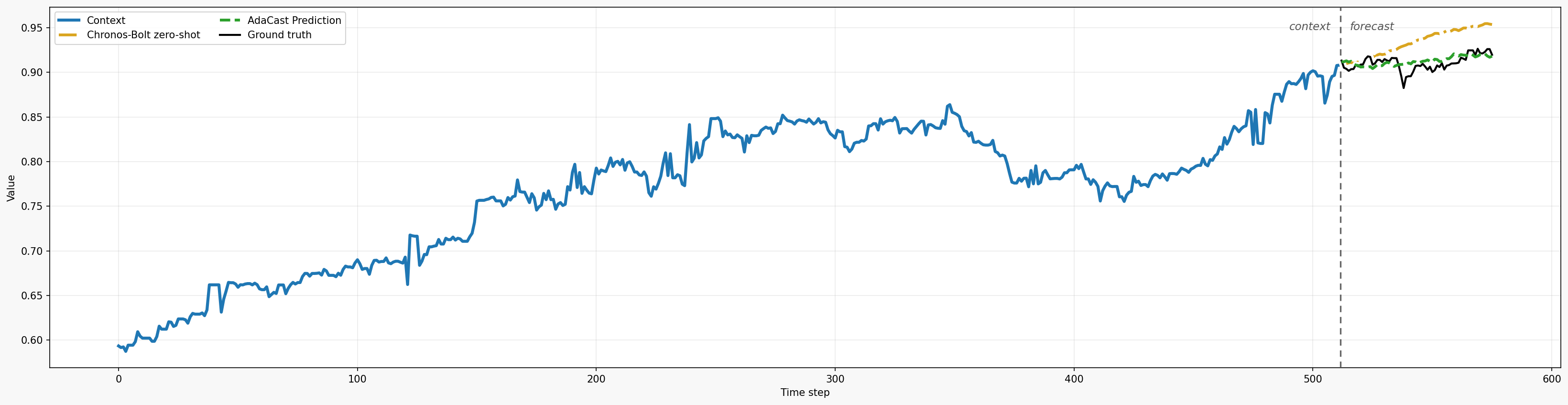}

    \vspace{-0.4em}
    {\small Exchange, Sample 2}
\end{minipage}

\vspace{0.5em}

\noindent
\begin{minipage}{\columnwidth}
    \centering
    \includegraphics[
        width=\linewidth,
        height=0.19\textheight,
        keepaspectratio
    ]{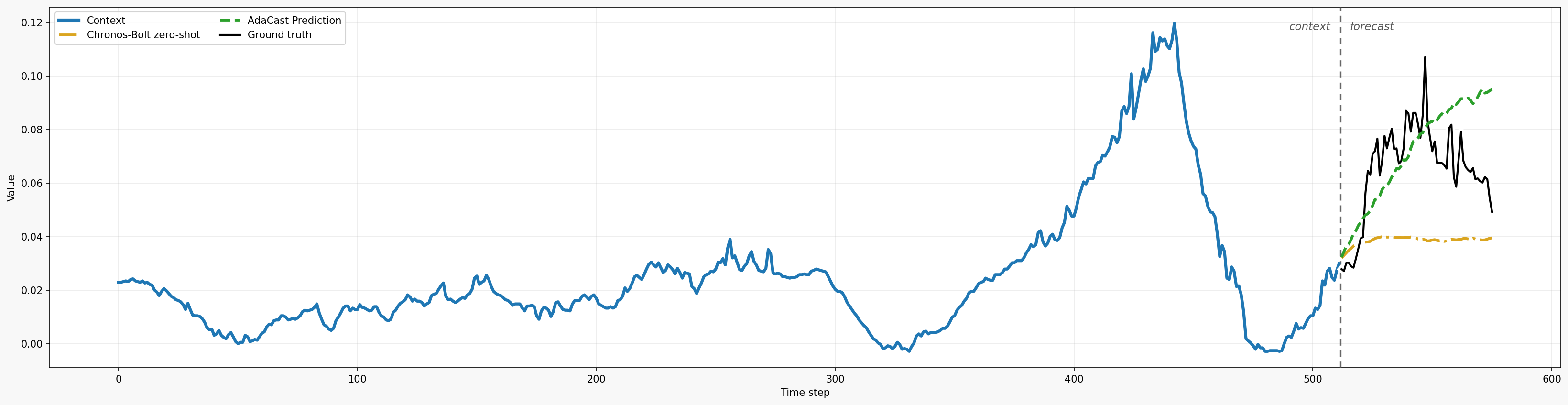}

    \vspace{-0.4em}
    {\small Weather, Sample 1}
\end{minipage}

\vspace{0.5em}

\noindent
\begin{minipage}{\columnwidth}
    \centering
    \includegraphics[
        width=\linewidth,
        height=0.19\textheight,
        keepaspectratio
    ]{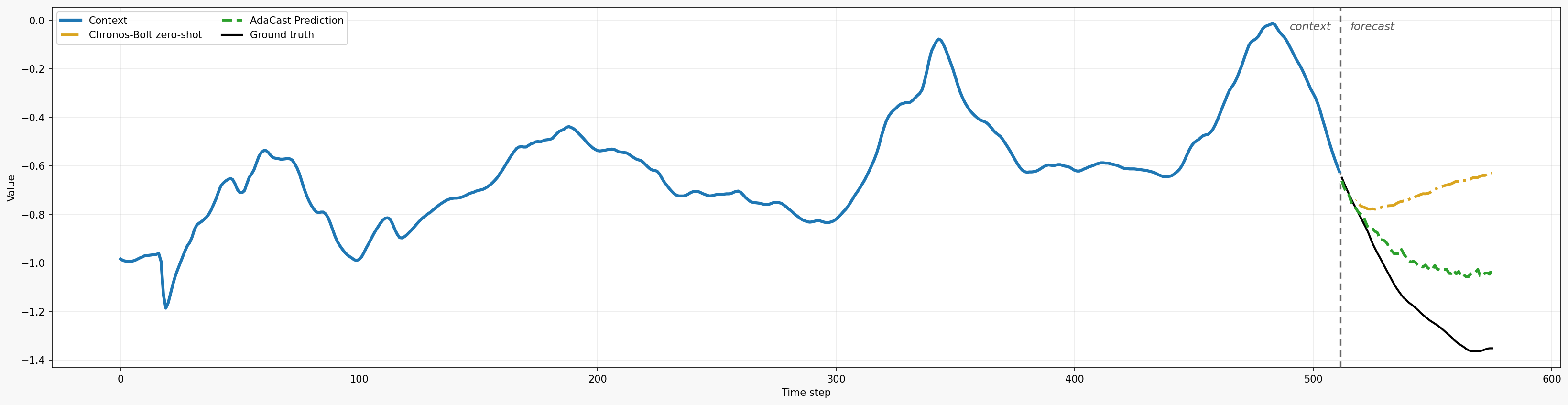}

    \vspace{-0.4em}
    {\small Weather, Sample 2}
\end{minipage}

\vspace{0.5em}

\end{document}